\PassOptionsToPackage{table}{xcolor}

\documentclass{article}
\usepackage[table]{xcolor}

\usepackage{microtype}
\usepackage{graphicx}
\usepackage{subfigure}
\usepackage{booktabs} 
\usepackage{enumitem}
\usepackage{listings}
\usepackage{array}

\usepackage{hyperref}
\usepackage[utf8]{inputenc}
\usepackage{amsmath}
\newcommand{\E}{\mathbb{E}}
\newcommand{\bP}{\mathbb{P}}

\usepackage{fancybox}
\usepackage{cancel}
     
\newcommand{\assi}{\textbf{Assistant: }}
\usepackage[skins]{tcolorbox} 
\tcbuselibrary{breakable} 
\newtcolorbox[auto counter, number within=section, list type=subsubsection, list inside=toc]{sectionbox}[2][]{
colback=white!98!gray, colframe=black, 
colbacktitle=white!90!gray, coltitle=black, 
fonttitle=\bfseries,
title={#2}, 
list entry={Comment \thetcbcounter\quad}
}

 \usepackage[accepted]{icml2025}

\usepackage{amssymb}
\usepackage{mathtools}
\usepackage{amsthm}

\newcommand{\KL}{D_{\mathrm{KL}}}
\newcommand{\cA}{\mathcal{A}}

\newcommand{\cD}{\mathcal{D}}

\newcommand{\cX}{\mathcal{X}}

\newcommand{\reff}{\mathrm{ref}}

\usepackage[capitalize,noabbrev]{cleveref}

\theoremstyle{plain}
\newtheorem{theorem}{Theorem}[section]
\newtheorem{proposition}[theorem]{Proposition}

\theoremstyle{definition}

\theoremstyle{remark}
\newtheorem{remark}[theorem]{Remark}

\usepackage[textsize=tiny]{todonotes}

\icmltitlerunning{Self-rewarding correction for mathematical reasoning}

\begin{document}

\twocolumn[
\icmltitle{Self-rewarding correction for mathematical reasoning}



\icmlsetsymbol{equal}{*}

\begin{icmlauthorlist}
\icmlauthor{Wei Xiong}{equal,yyy}
\icmlauthor{Hanning Zhang}{equal,yyy}
\icmlauthor{Chenlu Ye}{equal,yyy}
\icmlauthor{Lichang Chen}{sch}
\icmlauthor{Nan Jiang}{yyy}
\icmlauthor{Tong Zhang}{yyy}
\end{icmlauthorlist}

\icmlaffiliation{yyy}{University of Illinois Urbana-Champaign}
\icmlaffiliation{sch}{University of Maryland, College Park}

\icmlcorrespondingauthor{Wei Xiong}{wx13@illinois.edu}

\icmlkeywords{Machine Learning, ICML}

\vskip 0.3in
]



\printAffiliationsAndNotice{\icmlEqualContribution} 

\begin{abstract}
We study self-rewarding reasoning large language models (LLMs), which can simultaneously generate step-by-step reasoning and evaluate the correctness of their outputs during the inference time-\textit{without external feedback}. This integrated approach allows a single model to independently guide its reasoning process, offering computational advantages for model deployment.

We particularly focus on the representative task of self-correction, where models autonomously detect errors in their responses, revise outputs, and decide when to terminate iterative refinement loops. To enable this, we propose a two-staged algorithmic framework for constructing self-rewarding reasoning models using only self-generated data. In the first stage, we employ sequential rejection sampling to synthesize long chain-of-thought trajectories that incorporate both self-rewarding and self-correction mechanisms. Fine-tuning models on these curated data allows them to learn the patterns of self-rewarding and self-correction. In the second stage, we further enhance the models' ability to assess response accuracy and refine outputs through \textit{reinforcement learning} with rule-based signals. Experiments with Llama-3 and Qwen-2.5 demonstrate that our approach surpasses intrinsic self-correction capabilities and achieves performance comparable to systems that rely on external reward models.



\end{abstract}

\section{Introduction}
\label{sec:Intro}

Large language models (LLMs) have demonstrated remarkable capabilities in reasoning-related tasks such as mathematics and coding. Notable examples include ChatGPT \citep{OpenAI2023GPT4TR}, Claude \citep{Anthropic@claude}, and Gemini \citep{team2023gemini}. Following the release of GPT4-o1, LLMs with strong reasoning abilities have attracted even more attention, along with inference methods that enhance reasoning. A particularly desirable property of such models is their ability to detect inconsistencies and errors in self-generated responses—based on feedback to their prior outputs—and correct these errors to produce improved responses. This process is often referred to as \textit{self-correction} in the literature \citep{welleck2022generating, madaan2024self, kim2024language}.

When an external ground-truth reward model is available, studies \citep{kim2024language, qu2024recursive, shinn2024reflexion} have shown that LLMs can refine their initial responses based on external gold reward feedback and determine when to terminate the self-correction loop. These approaches have proven effective for both mathematical reasoning and general agent tasks. Moreover, even when relying on imperfect proxy rewards, models can still achieve higher accuracy in revised responses by leveraging feedback from an outcome-based reward model (see Section~\ref{sec:exp} for empirical results). However, since these reward models are often themselves LLMs, deploying them requires running multiple models during inference, which increases computational costs and deployment complexity. In contrast, without external reward feedback, current LLMs struggle to refine their initial responses solely based on their intrinsic capabilities—a limitation known as intrinsic self-correction \citep{huang2023large}.

While reward models are traditionally trained with an additional scalar head for general-purpose chat \citep{ouyang2022training, bai2022training, touvron2023llama} and reasoning tasks \citep{cobbe2021gsm8k, lightman2023let}, recent work suggests that LLMs themselves can generate reward signals in a generative way. For example, the LLM-as-a-judge approach \citep{zheng2023judging, dubois2023alpacafarm} prompts the LLM to evaluate text outputs, effectively serving as a surrogate for human feedback. Another emerging direction explores generative reward models \citep{zhao2023slic, dong2024rlhf, zhang2024generative, mahan2024generative, zhang2024entropyregularizedprocessrewardmodel}, which formulate evaluation tasks as instruction-following problems, using the probability of generating specific tokens as the reward value. These methods leverage LLMs' next-token prediction capabilities, integrate the generation and evaluation into a unified framework.

Building on these insights, this work investigates self-rewarding reasoning models that can incorporate three abilities within a single LLM: (i) generating step-by-step reasoning paths for given prompts, (ii) evaluating the correctness of generated responses, and (iii) revising and enhancing previous responses based on self-rewarding signals. Our key contributions are as follows:
\begin{enumerate}
    \item \textbf{Self-rewarding reasoning framework.} We introduce a self-rewarding reasoning framework for LLMs, which integrates the generator and reward model into a single LLM, enabling autonomous reasoning, evaluation, and correction. This unification simplifies the model’s decision-making process and reduces computational overhead compared to external reward-based approaches.
    \item \textbf{Algorithmic framework for self-correction.} We focus on the self-correction in mathematical reasoning and propose a two-stage framework that relies only on self-generated data. In the first stage, we use sequential rejection sampling to construct long chain-of-thought (CoT) trajectories that encode both self-rewarding and self-correction behaviors. Fine-tuning models on these trajectories enables them to detect the error in the self-generated responses and revise the previous attempts. In the second stage, we further enhance these patterns through reinforcement learning with rule-based signals. 
    \item \textbf{Empirical validation and analysis.} Through extensive experiments, we show that self-rewarding correction significantly outperforms intrinsic self-correction. Additionally, we conduct ablation studies to investigate the learning dynamics of the proposed framework, providing deeper insights into its behavior and effectiveness. The training codes and datasets are publicly available on GitHub\footnote{\url{https://github.com/RLHFlow/Self-rewarding-reasoning-LLM}}.
\end{enumerate}

\section{Related Work}
We review the works that are mostly related to our project in this section.

\paragraph{Self-rewarding alignment.} Our work aligns with research on self-rewarding alignment \citep{yuan2024self, prasad2024self}, where both of our project and their methods share similar spirits that we can unify the generation ability and evaluation ability into a single LLM. These methods leverage iterative DPO-type algorithms, where the model labels its own generated responses to provide training signals for subsequent iterations, enabling \textit{self-improvement}. In contrast, our approach does not focus on self-improvement during training. Instead, we rely on an external ground-truth reward model to provide learning signals in training. Our study emphasizes inference-time alignment for reasoning-focused LLMs, where self-rewarding signals are employed solely to guide inference rather than training.

\paragraph{Self-correction.} Our work is closely related to self-correction in LLMs. We refer interested readers to the survey \citep{pan2023automatically} for a more comprehensive review and only review some representative approaches that are mostly related to our project. \citet{li-etal-2024-selective} demonstrated that incorporating teacher model reflections into SFT data enhances students' self-reflection abilities in general-purpose conversation tasks. However, for reasoning tasks, \citet{huang2023large} found that current LLMs—without additional training—fail to self-correct purely through intrinsic reasoning (i.e., prompting). This observation is also validated in \citet{qu2024recursive, tyen2023llms, zheng2024natural}. A more in-depth analysis shows that most prior successful studies in this domain depend on \textbf{external} (ground-truth) reward models to determine when to initiate and terminate self-correction \citep{kim2024language, qu2024recursive, shinn2024reflexion, madaan2024self}. Currently, there is no major work demonstrating that intrinsic self-correction (via prompting or fine-tuning) is reliably effective. Furthermore, because external reward models are typically LLM-based, these methods introduce additional computational overhead by requiring a multi-agent system for inference.

Recognizing this challenge, our study explores how LLMs can autonomously evaluate response quality and correct errors without external reward models. Specifically, we introduce a self-rewarding reasoning framework that enables a single LLM to perform error detection and self-correction effectively. Among the works in self-correction, the most relevant work is the recent \citet{kumar2024training}, which employed a multi-turn deep RL approach to train self-correcting models. In comparison, this work introduces a new and general self-rewarding formulation for reasoning-focused LLMs, with self-correction as a representative application. Compared to the intrinsic correction and the framework in \citet{kumar2024training}, one major difference is that our framework equips models with self-rewarding ability, enabling our models to intelligently scale inference compute by \textit{selectively} revising the first attempts, which helps to reduce computational overhead by avoiding unnecessary iterations. We will also design experiments to illustrate this idea. 

Algorithmically, our approach also differs from \citet{kumar2024training}. We first use sequential rejection sampling to construct long CoT trajectories with both self-rewarding and self-correction patterns, which serve as warm-up fine-tuning data. We then enhance these behaviors through reinforcement learning (using either DPO-type algorithms or PPO) with rule-based signals. In contrast,  \citet{kumar2024training} employed RLOO \citep{ahmadian2024back} with a specialized reward function for a two-turn self-correction task. While their no-public models (Gemini) and implementation details (parameters, codes) do not enable comparison, we believe that the multi-turn RL methods proposed by \citet{kumar2024training} could also complement the proposed self-rewarding framework, and achieve better reasoning performance compared to the standard reasoning models.
\vspace{-8pt}
\paragraph{Rule-based RL for LLMs mathematical reasoning.} Rule-based reinforcement learning has received significant attention following the success of DeepSeek-R1 \citep{deepseekai2025deepseekr1incentivizingreasoningcapability}. Open-source efforts have since attempted to replicate its performance using Qwen models \citep{yang2024qwen2}, including works such as \citet{zeng2025simplerl, cui2025processreinforcementimplicitrewards, zhang2025dpor1}. These methods train LLMs using only the correctness score (whether the final answer is correct or not) and a format score (whether the final answer is output in a pre-determined format), in contrast to the previous works with the neural network-based reward model \citep{cobbe2021gsm8k, lightman2023let, zhang2024entropyregularizedprocessrewardmodel}.  
In particular, \citet{deepseekai2025deepseekr1incentivizingreasoningcapability} observed that self-correction naturally emerges during RL training (referred to as an AHA moment in their report). However, our preliminary experiments, along with open-source replications using Qwen-2.5-Math \citep{liu2025oatzero, zhang2025dpor1, cheng2025pure}, suggest that (i) the base models already exhibit some self-correction ability, though it is quite sparse. (ii) vanilla rule-based RL \textbf{cannot} consistently enhance self-correction without additional design.

Interestingly, even when using the same algorithms and data, similar improvements in mathematical reasoning are not observed in models such as Llama \citep{meta_llama3, touvron2023llama}. We hypothesize that Qwen-2.5-Math and DeepSeek-R1 benefit from extensive pre-training on high-quality mathematical corpora (e.g., 1T tokens for Qwen-2.5-Math \citep{yang2024qwen2}), and that the AHA moment may stem from carefully curated data containing self-correction patterns in pre-training or a cool-down stage. Since these datasets are non-public, the exact details remain unknown.

In contrast, our study shows that a warm-up stage using a carefully curated SFT dataset (collected via sequential rejection sampling) enables models to learn self-correction patterns more reliably. This foundation allows rule-based RL to further enhance these behaviors in a stable manner. We also remark that our two-stage framework and most of the associated experiments are performed prior to the release of DeepSeek-R1.

\begin{table*}[t]
\centering
\begin{minipage}{2.0\columnwidth}\vspace{0mm}    \centering
\begin{sectionbox}[]
    \centering
      \small
\begin{tabular}{p{0.97\columnwidth} c}
    \textbf{User:} A particular star has an absolute magnitude $M=-7$. If this star is observed in a galaxy that is at a distance of $3 \mathrm{Mpc}$, what will its apparent magnitude be? \\ Please reason step by step, and put your final answer within $\boxed{}$. \textcolor{blue}{(prompt $x$)}\\\\

    \textbf{Assistant:} To determine the apparent magnitude of the star, we can use the distance modulus formula : $ m - M = 5 \log_{10}(d) - 5 $ where:$ m $ is the apparent magnitude,$ M $ is the absolute magnitude,$ d $ is the distance to the star in parsecs.... The apparent magnitude of the star, when observed from a distance of 3 Mpc, is approximately $ 7.58 $. So the final answer is: $\boxed{7.58}$. \textcolor{blue}{(initial response $a^1$)}\\\\
    To verify the correctness of the solution, let's recheck the steps ... So the final answer is: $\boxed{20.39}$. The self-evaluation is: [VERIFY] wrong. \textcolor{blue}{(self-rewarding evaluation $y^1$)} \\\\
    
    Let's correct it step by step. Given: Absolute magnitude ... The correct apparent magnitude is approximately $ 20.39 $. So the final answer is: $\boxed{20.39}$. \\\\ \textcolor{blue}{(revised response $a^2$)}.

    Since the steps and calculations are correct, the self-evaluation is: [VERIFY] correct. \textcolor{blue}{(self-rewarding evaluation $y^2$)}
    \end{tabular}
\end{sectionbox}
\caption{An example of the self-rewarding reasoning path. We omit the detailed reasoning path for a clear presentation. The full trajectory is available at Table~\ref{tab:example_hh_rlhf_dataset} in Appendix.} 
    \label{fig:example0}
\end{minipage}
\end{table*}

\section{Self-rewarding Reasoning Language Models} 

We formulate the self-rewarding reasoning process as a multi-turn Markov Decision Process (MDP). After observing an initial prompt $s^1=x \in \cX$ from some distribution $d_0$, an LLM, denoted as $\pi$, will generate an initial reasoning attempt $a^1 \sim \pi^1(\cdot|s^1)$ from the action space $\cA$. The LLM then self-rewards its response by generating an evaluation:
$$
y^1 \sim \pi^1(\cdot|s^1,a^1).
$$
If the model assesses its answer as correct ($y^1=\text{[VERIFY] correct}$, details provided later), the generation stops. Otherwise, the LLM proceeds to the next step, generating a refined response and evaluation:
$$(a^2,y^2)\sim\pi^2(\cdot|s^2),$$
where the generation is conditioned on the updated state $s^2=(s^1,a^1,y^1)$. The self-refinement process continues until the model produces a self-evaluation $y^h$ that assesses the answer as correct. 

We assume that we have access to the ground-truth verifier $r^\star: \cX \times \cA \to \{0, 1\}$, which determines whether a response is correct. Throughout this study, we use the ToRA verification script \citep{gou2023tora}, built on the Python library SymPy for symbolic mathematics. We also present a representative Example~\ref{fig:example0} to illustrate the process. 

\paragraph{Two-stage training framework.} Following standard post-training practices for LLMs, we adopt a two-stage approach:
\begin{enumerate}
    \item \textbf{Self-rewarding instruction-following fine-tuning (IFT).} Starting with an initial LLM $\pi_0$ (e.g., a general-purpose chatbot), we collect demonstration data by a sequential rejection sampling process and fine-tune $\pi_0$ to get an improved model $\pi_{\reff}$, which integrates self-rewarding reasoning abilities.
    \item \textbf{Reinforcement learning (RL) optimization.} We further refine $\pi_{\reff}$ using RL, leveraging it as the reference model. This stage can further enhance the model's ability to assess correctness and refine previous responses.
\end{enumerate}

\subsection{Self-rewarding Instruction-following Fine-tuning} \label{sec:ift}
\paragraph{Self-rewarding by token prediction.} To train the LLMs to evaluate the reasoning steps, we formulate this task as an instruction-following task, following prior works \citep{zhao2023slic, dong2024rlhf, liu2023statistical, ye2024theoretical, wang2024math, zhang2024generative}. Specifically, we allow models to include reasoning in their evaluations while requiring them to output specific tokens to indicate their evaluation results. We experimented with different token choices, such as: (i) a prompt ``Is the most recent final answer correct (Yes or No)?'' with ``Yes'' and ``No'' as the response tokens, as used in \citep{xie2023decomposition, zhang2024generative}; (ii) explicit markers such as ``[VERIFY] correct'' and ``[VERIFY] wrong''. Our experiments show no significant performance differences between these choices. During inference, rather than using the likelihood of ``Yes'' as a reward (as in \citep{zhao2023slic, dong2024rlhf, zhang2024generative}), we sample the evaluation token from the distribution. This allows us to use a standard inference pipeline without any specific adjustment. See Table~\ref{fig:example0} for an example. 

\begin{remark}
We choose these specific tokens primarily for research simplicity. However, we expect that similar results can be achieved even if these special tokens are replaced with more natural language expressions, such as ``wait'', ``aha'', or ``let me re-check the answer'', where one can also leverage the LLMs to complete this paraphrasing process. 
\end{remark}

\begin{table*}[t]
\centering
\begin{minipage}{2.0\columnwidth}\vspace{0mm}    \centering
\begin{sectionbox}[]
    \centering
      \small
\begin{tabular}{p{0.97\columnwidth} c}
 \textbf{System:} You are a mathematical reasoning assistant. For each problem, follow these steps strictly: 1. Solve the problem using step-by-step reasoning and output the final answer within \boxed{}. Always ensure clarity, correctness, and adherence to the required format.\\\\
    
    \textbf{User:} How many 3-letter words can we make from the letters A, B, C, D, and E, if we are allowed to repeat letters, and we must use the letters A and B at least once each in every word? (Here, a word is an arbitrary sequence of letters.) Let's think step by step and output the final answer within \boxed{ }.\\\\

    \textbf{Assistant:} Here's how we can break it down: 1. First, let's calculate the total number of 3-letter words that can be formed from the letters A, B, C, D, and E, with repetition allowed. ... 7. Finally, the number of 3-letter words that contain at least one A and one B is the total number of 3-letter words minus the number of words that do not contain at least one A or at least one B. The number of 3-letter words that can be made from the letters A, B, C, D, and E, where each word must contain at least one A and one B, is $\boxed{38}$.\\\\ 

    \textbf{User:} Perform a self-evaluation: You may include reasoning to verify correctness. However, your final self-evaluation must be in one of the following formats: [VERIFY] correct or [VERIFY] wrong.\\\\

    \textbf{Assistant:} Let's perform a self-evaluation of the solution. We first calculate ... This is $125 - 101 = 24$. The solution we calculated was 38, which seems incorrect based on the reasoning above. The correct answer should be 24. {\color{red} Let's output the self-evaluation: [VERIFY] wrong.}\\\\

    \textbf{User:} please identify the mistake in your previous reasoning, revise your reasoning path and output a corrected final answer within \boxed{}\\\\
    
    \textbf{Assistant:} Let's correct the mistake in the previous reasoning. The correct steps should be as follows: 1. Calculate ... 6. Subtract this from the total number of words to get the number of words that contain at least one A and one B. This is $125 - 101 = 24$. The correct answer is indeed $\boxed{24}$.
    \end{tabular}
\end{sectionbox}
\caption{An example of the sequential rejection sampling to collect long CoT trajectories.} 
    \label{fig:example_2}
\end{minipage}
\end{table*}

\textbf{Data collection by sequential rejection sampling.} We employ a rejection sampling approach, similar to STaR \citep{zelikman2022star} and RAFT \citep{dong2023raft}, where we generate a large amount of self-correction trajectories and only preserve the desired trajectories. The major difference is that since the self-correction behavior is sparse in base models and self-rewarding pattern is missing, it is unlikely to collect the desired trajectory directly. In view of this, we sequentially prompt the base model and generate different steps separately. Then, we combine them into long CoT trajectories that incorporate both self-rewarding and self-correction patterns. 

Our data collection process consists of the following steps:
\begin{enumerate}
    \item \textbf{Generating initial reasoning responses}: training prompts from datasets such as MATH \citep{hendrycks2021measuring} and GSM8K \citep{cobbe2021gsm8k} and sample $N_1=50$ initial responses $a^1$ per prompt as our base trajectories (see Section~\ref{sec:exp} for details of experiment setups).
    \item \textbf{Self-rewarding signal sampling:} For each prompt and initial response, we further sample $N_2=8$ self-evaluations and keep only one evaluation result that is the same as the ground truth. Then, we split them into $G^{\mathrm{correct}}$ and $G^{\mathrm{wrong}}$ using the ground-truth verifier $r^\star$.
    \item \textbf{Correction sampling}: For each prompt and initial response in $G^{\mathrm{wrong}}$, we sample $M_1=8$ completions by providing the feedback that the initial response was wrong to collect trajectories that successfully revise incorrect responses. For each prompt and initial response in $G^{\mathrm{correct}}$, however, we also tell the model that the response was incorrect and collect $M_2=4$ completions. By doing so, we want to additionally collect ``correct-to-correct'' trajectories in the face of wrong judgment.
\end{enumerate}
Eventually, we collect $8 \times |G^{\mathrm{wrong}}| + 4 \times |G^{\mathrm{correct}}|$ full trajectories. Then, we filter the dataset and only keep the following types of data:
\begin{itemize}[itemsep=2pt]
    \item $\mathcal{D}^{\mathrm{IFT}}_1:$ wrong $a^1$, $y^1=\text{[VERIFY] wrong}$, correct $a^2$;
    \item $\mathcal{D}^{\mathrm{IFT}}_2:$ correct $a^1$, $y^1=\text{[VERIFY] wrong}$, correct $a^2$;
    \item $\mathcal{D}^{\mathrm{IFT}}_3:$ correct $a^1$, $y^1=\text{[VERIFY] correct}$. 
\end{itemize}
We provide an example of data collection process in Table~\ref{fig:example_2}. We limit the horizon to two iterations due to resource constraint, and preserve at most one trajectory per base sample to control dataset size. Then we fine-tune the LLMs using standard SFT pipeline to maximize:
\begin{equation} \label{eqn:sft_target}
\footnotesize
    \begin{aligned}
         \sum_{\mathcal{D}^{\mathrm{IFT}}_1}\big[ \log P(y^1|x,a^1) + \log P(a^2|x,a^1,y^1) \big] \\
        + \sum_{\mathcal{D}^{\mathrm{IFT}}_2} \log P(a^2|x,a^1,y^1) + \sum_{\mathcal{D}^{\mathrm{IFT}}_3} \log P(y^1|x,a^1).
    \end{aligned}
\end{equation}

In practice, however, we observe that the multi-task training can lead to stability issue and can slightly hurt the first-round performance. To mitigate this issue, we also train on the correct attempt $a^1$ for the samples in $\mathcal{D}^{\mathrm{IFT}}_3$.

\subsection{KL-regularized Reinforcement Learning}
\label{sec:dpo}
In this stage, we aim to further enhance the self-rewarding IFT models using reinforcement learning. We consider both deep RL methods \citep{schulman2017proximal} and direct alignment algorithms \citep{zhao2023slic, rafailov2023direct, azar2023general, liu2023statistical}.

\paragraph{Learning signal.} To facilitate the reinforcement learning stage, we assume there exists a trajectory-wise reward function $u^\star(\tau)$ for trajectory 
$$\tau=(x,a^1,y^1,\ldots,a^H,y^H).$$ 
However, instead of learning a proxy reward from data like the BT model in RLHF \citep{ouyang2022training} or outcome-supervised reward (ORM) in previous mathematical reasoning literature \citep{lightman2023let}, we primarily use the oracle reward 
$$u^\star(\tau) = r^\star(x, a^H),$$
i.e., whether the final result is correct or not. The main advantage is that the oracle reward can largely mitigate the risk of reward hacking. This is also referred to as the \textit{rule-based RL} in the very recent literature \citep{deepseekai2025deepseekr1incentivizingreasoningcapability}. We will also study the additional rule designs for either reward value assignment (PPO training) or data ranking (DPO training), where an implicit $u^\star$ is determined by the set of rules we use.

Following standard RLHF methodologies \citep{ouyang2022training, bai2022training}, we optimize the following KL-regularized objective:
\begin{equation} \label{eqn:target}
{\footnotesize
\begin{aligned}
    &\max_{\pi \in\Pi}\E_{x \sim d_0, a^1 \sim\pi^0(\cdot|x)}\E_{\tau\sim\pi(\cdot|x,a^1)}  \Big[\\    
    &\qquad u^\star(\tau) - \eta\sum_{h=1}^H \KL(\pi^h(\cdot|s^h),\pi^h_{\reff}(\cdot|s^h))\Big].
\end{aligned}
}
\end{equation}
The optimal policy, as well as its associated optimal value satisfies the following optimality condition \citep{xiong2024building, xie2024exploratory, zhong2024dpo}.
\begin{proposition} \label{prop:optimality}
    We can recursively define the following optimal value functions and optimal policies for a KL-regularized MDP with horizon $H$ and deterministic external observation. For Q value, we have
    \begin{equation}
    \begin{aligned}
                Q^\star_h(s^h, a^h,y^h) =\begin{cases}
    & u^\star(s^H, a^H,y^H),  \qquad \text{ if } h = H,  \\
  & V^\star_{h+1}(s_{h+1}), \qquad \text{ if } h \leq H-1.
\end{cases}   
    \end{aligned}
    \end{equation}
    Also, for all $h \in [H]$, we have:
    \begin{equation*}
    {\footnotesize
        \begin{aligned}
            V_\star^h(s^h)  = \eta \log \underbrace{\mathbb{E}_{a^h,y^h \sim \pi_{\mathrm{ref}}^h(\cdot\mid s^h)} \exp\Big(\frac{Q_\star^h(s^h,a^h,y^h)}{\eta}\Big)}_{=: Z^h(s^h)},
        \end{aligned}}
    \end{equation*}
    \begin{equation} \label{eqn:3}
    {\footnotesize
        \begin{aligned}
            \pi_\star^h(a^h,y^h \mid s^h) = \frac{\pi^h_{\mathrm{ref}}(a^h, y^h\mid s^h)}{Z^h(s^h)} \cdot \exp \Big(\frac{Q_\star^h(s^h, a^h, y^h)}{\eta}\Big).
        \end{aligned}}
    \end{equation} 
\end{proposition}
We remark that one advantage of the proposition is that it allows deterministic external message (e.g. instruction prompts) in the state update, which will be useful when we consider a simplified research framework in Section~\ref{sec:exp}.

We also adopt Direct Preference Optimization (DPO) \citep{rafailov2023direct, azar2023general, zhao2023slic, ethayarajh2024kto} to solve Equation~\ref{eqn:target}, primarily due to computational constraints. In particular, we use the multi-turn DPO (M-DPO) framework from \citet{xiong2024building}, since it allows deterministic external observation in the state transition. To facilitate direct preference learning and bypass explicit reward training, we impose the following trajectory-level Bradley-Terry (BT) preference structure \citep{bradley1952rank}. Specifically, given two trajectories $\tau^1, \tau^2$, the probability of $\tau^1$ being preferred than $\tau^2$, denoted as $\tau^1\succ\tau^2$, is
    \begin{equation*}
    \begin{aligned}
        \bP(\tau^1\succ\tau^2\,|\, \tau^1,\tau^2) = \sigma(u^\star(\tau^1)-u^\star(\tau^2)),
    \end{aligned}
    \end{equation*}
    where $\sigma(z)=1/(1+\exp(-z))$ is the sigmoid function. Following \citet{xiong2024building}, we take log on both sides of \eqref{eqn:3}, and connect a utility function $u_\theta$ with associated policy $\pi_\theta$ and value $V_\theta$:
\begin{equation*}
\footnotesize
\begin{aligned}
    &\log\frac{\pi_\theta^h(y^1|s^1)}{\pi_\reff^1(y^1|s^1)}= V_\theta^2(s^2) - V_\theta^1(s^1),\\
    &\log\frac{\pi_\theta^h(a^h,y^h|s^h)}{\pi_\reff^h(a^h,y^h|s^h)} = Q_\theta^h(s^h,a^h,y^h) - V_\theta^h(s^h).
\end{aligned}
\end{equation*}
For a pair of trajectories $\tau^w,~\tau^l$ where $\tau^w\succ\tau^l$, we have
\begin{equation*}
\footnotesize
\begin{aligned}
    &u_\theta(\tau^w) - u_\theta(\tau^l) = \log\frac{\pi_\theta^1(y_w^1|x,a^1)}{\pi_\reff^1(y_w^1|x,a^1)} -\log\frac{\pi_\theta^1(y_l|x,a^1)}{\pi_\reff^1(y_l|x,a^1)}\\
    &\qquad + \sum_{h=1}^H \log\frac{\pi_\theta^h(a_w^h,y_w^h|s_w^h)}{\pi_\reff^h(a_w^h|s_w^h)} - \log\frac{\pi_\theta^h(a_l^h,y_l^h|s_l^h)}{\pi_\reff^h(a_l^h,y_l^h|s_l^h)}.
\end{aligned}
\end{equation*}
Taking this reward difference parameterization into the log-likelihood of the BT model $
   \sum_{(\tau^w, \tau^l) \in \cD} \log \sigma\big(u_\theta(\tau^w) - u_\theta(\tau^l)\big),$
we obtain the loss function $\mathcal{L}_{\textup{M-DPO}}(\theta)$:
\begin{equation}
    \label{eqn:m_dpo_loss} 
    \footnotesize
    \begin{aligned}
             &-\sum_{(\tau^w, \tau^l) \in \cD} \log \sigma \Big( 
            \eta  \Big[\log\frac{\pi_\theta^1(y_w^1|x,a^1)}{\pi_\reff^1(y_w^1|x,a^1)} -\log\frac{\pi_\theta^1(y_l|x,a^1)}{\pi_\reff^1(y_l|x,a^1)}\\
    &\qquad + \sum_{h=1}^H \log\frac{\pi_\theta^h(a_w^h,y_w^h|s_w^h)}{\pi_\reff^h(a_w^h|s_w^h)} - \log\frac{\pi_\theta^h(a_l^h,y_l^h|s_l^h)}{\pi_\reff^h(a_l^h,y_l^h|s_l^h)}\Big] \Big).
                \end{aligned}
\end{equation} 

\begin{table*}[h]
    \centering \footnotesize
        \caption{Main results of experiments with Qwen2.5-Math-7B-base. The single-turn baselines are used to train a regular CoT reasoning model. 
        The baselines with $^\dagger$ perform self-correction under the external prompt, where training may apply to enhance this ability. We use greedy decoding following the convention of the recent open-source projects on mathematical reasoning.} \vspace{10pt} \label{tab:scaled_result}
    \begin{tabular}{cc|ccccc}
    \toprule
\textbf{Benchmark} & \textbf{Method}  & \textbf{Turn 1} & \textcolor{red}{\textbf{Final Accuracy}}& \textcolor{red}{$\Delta(t_1,t_2)$} & $\Delta^{i \to c}(t_1,t_2)$ & $\Delta^{c \to i}(t_1,t_2)$ \\ 
    \midrule 
    &Single-turn STaR/RAFT & 77.0 & 77.0& - & - & - \\
  &  Single-turn DPO & 76.8 & 76.8 & - & - & - \\
  &  Single-turn PPO & 79.4 & 79.4 & - & - & - \\
   & Prompt with Gold RM$^\dagger$ & 65.4 & 66.8 & 1.4 & 1.4 & 0.0 \\
   & Intrinsic self-correction$^\dagger$ & 65.4 & 51.4 & -14.0 & 1.4 & 15.4 \\
 MATH500&   STaR/RAFT for self-correction$^\dagger$ & 71.6 & 70.4 & -1.2 & 5.0 & 6.2 \\
   & STaR/RAFT+ for self-correction$^\dagger$ & 72.0 & 71.2 & -0.8 & 3.0 & 3.8 \\
    \rowcolor[rgb]{ .867, .922, .969} &Self-rewarding IFT & 72.6 & 77.2 & 4.6 & 5.0 & 0.4 \\
    \rowcolor[rgb]{ .867, .922, .969} &Self-rewarding IFT + DPO w correctness & 72.8 & 78.6 & 5.8 & 6.0 & 0.2 \\
    \rowcolor[rgb]{ .867, .922, .969} &Self-rewarding IFT + PPO w correctness & 75.8 & 80.2 & 4.4 & 4.8 & 0.4 \\
    \midrule 
  &  Single-turn STaR/RAFT & 40.1 & 40.1 & - & - & - \\
   & Single-turn DPO & 39.0 & 39.0 & - & - & - \\
   & Single-turn PPO & 39.5 & 39.5 & - & - & - \\
   & Prompt with Gold RM$^\dagger$ & 23.4 & 25.6 & 2.2 & 2.2 & 0 \\
   & Intrinsic self-correction$^\dagger$ & 23.4 & 18.1 & -5.3 & 2.2 & 7.5 \\
   OlympiadBench & STaR/RAFT for self-correction $^\dagger$& 36.5 & 32.5 & -4.0 & 7.2 & 11.2  \\
    &STaR/RAFT+ for self-correction$^\dagger$ & 35.7 & 35.5 & -0.2 & 3.2 & 3.4 \\
    \rowcolor[rgb]{ .867, .922, .969} &Self-rewarding IFT & 35.4 & 39.4 & 4.0 & 4.7 & 0.7\\
    \rowcolor[rgb]{ .867, .922, .969} &Self-rewarding IFT + DPO w correctness & 37.6 & 40.1 & 2.5 & 3.5 & 1.0\\
    \rowcolor[rgb]{ .867, .922, .969} &Self-rewarding IFT + PPO w correctness & 41.0 & 43.4 & 2.4 & 2.8 & 0.4\\
       \midrule 
    &Single-turn STaR/RAFT & 32.0 &32.0& - & - & - \\
    &Single-turn DPO & 31.6 & 31.6 &- & - & - \\
   & Single-turn PPO & 33.1 & 33.1 & - & - & - \\
   & Prompt with Gold RM$^\dagger$ & 9.9 & 11.7 & 1.8 & 1.8 & 0\\
    &Intrinsic self-correction$^\dagger$ & 9.9 & 8.4 & -1.5 & 1.8 & 3.3 \\
  Minerva Math &  STaR/RAFT for self-correction$^\dagger$ & 28.7 & 29.4 & 0.7 & 1.8 & 1.1  \\
  &  STaR/RAFT+ for self-correction$^\dagger$ & 25.7 & 25.3 & -0.4 & 0.8 & 1.2 \\
    \rowcolor[rgb]{ .867, .922, .969} &Self-rewarding IFT & 23.2 & 28.7 & 5.5 & 7.3 & 1.8 \\
    \rowcolor[rgb]{ .867, .922, .969} &Self-rewarding IFT + DPO w correctness & 26.8 & 34.6 & 7.8 & 9.6 & 1.8\\
    \rowcolor[rgb]{ .867, .922, .969} &Self-rewarding IFT + PPO w correctness & 34.0 & 38.4 & 4.4 & 5.1 & 0.7\\ 
    \bottomrule
    \end{tabular}
    \end{table*}
\section{Experiment Results}

\paragraph{Task, datasets, and data format.} We evaluate models' mathematical reasoning abilities using standard benchmarks, including MATH500 \citep{hendrycks2020measuring}, OlympiadBench \citep{he2024olympiadbench}, and Minerva Math \citep{lewkowycz2022solving}. These datasets provide a moderate size for reliable and efficient model evaluation, covering topics such as algebra, geometry, probability, number theory, and calculus. For training, we mainly use the prompts in NumiaMath-CoT dataset \citep{numina_math_7b}. Specifically, we use a 50K subset for the self-rewarding IFT stage, a 10K subset for validation and model selection, and the remaining data for RL training. During inference, the model generates up to 4096 tokens, with VLLM 0.5.4 \citep{kwon2023efficient} accelerating the process. 

\textbf{Evaluation metrics.} We employ two categories of metrics to evaluate our models: (1) mathematical reasoning and self-correction and (2) reward model accuracy. First, we follow \citet{kumar2024training} to consider the following metrics to evaluate the models' ability of mathematical reasoning and self-correction.
\begin{enumerate}
\setlength{\itemsep}{1pt}
    \item \textbf{Turn 1}: accuracy of the first attempt;
    \item \textbf{Final accuracy}: accuracy of the final answer;
    \item $\Delta(t_1, t_2)$: improvement in accuracy from the first attempt to the final answer;
    \item $\Delta^{i \to c}(t_1,t_2)$: fraction of problems changed from incorrect to correct;
    \item $\Delta^{c \to i}(t_1,t_2)$: fraction of problems changed from correct to incorrect.
\end{enumerate}
Due to the nature of the self-rewarding reasoning framework, we additionally include the metrics to measure the accuracy as a reward model. We also defer a more comprehensive understanding of the proposed framework with a slightly simplified template to next section, where we will additionally compute the ratio of modifying a correct answer to incorrect when facing a misleading reward.
\begin{enumerate}
    \item   \textbf{RM Accuracy} $(a, b)$: class-dependent accuracy for correct and incorrect trajectories. In other words, $a$ is the true positive rate and $b$ is the true negative rate;
    \item \textbf{Ratio} $p^{c\to i}(t_1, t_2)$: probability of modifying a correct answer to incorrect when facing a misleading reward.
\end{enumerate}
For all evaluations, we use zero-shot CoT prompting and greedy decoding following the convention of recent projects with Qwen-2.5-Math models.

\textbf{Experiment setup of self-rewarding IFT.} We use Qwen2.5-Math-7B-base as the base model, which is continuously pre-trained on extensive mathematical and instruction-following data. Sequential rejection sampling (introduced in Section~\ref{sec:ift}) is used for data collection, resulting in a dataset of 32K trajectories, where we roughly balance between correct and incorrect first attempts. In fine-tuning, samples are packed into 8192-token blocks and we use a learning rate of 1e-5, a cosine scheduler, and a 0.05 warm-up ratio. Global batch size is set to be 32. We train the models for three epochs and eventually select the one at the end of the first epoch.

\textbf{Experiment setup of reinforcement learning.} For iterative DPO training, we adopt setups from \citet{xiong2024building} with a learning rate of $2\times 10^{-7}$, a cosine scheduler, and a batch size of 32. We tune $\eta \in \{0.1, 0.5\}$ and also train with and without an NLL loss in the DPO objective \citep{pang2024iterative, xie2024exploratory, liu2024provably}. For each iteration, we use 20K prompts and collect $8$ responses per prompt. Then, we extract the comparison pairs using the correctness score. If all responses admit the same score, we skip the prompt. A 10K validation set from NuminaMath-CoT is used for model selection. The primary metric for model selection is accuracy at turn 2. When models achieve comparable turn-2 accuracy, we choose the models with higher $\Delta(t_1,t_2)$ improvement. The best model of these training setups is used as the representative model. For PPO training, we mainly follow a pulic example script of veRL \citep{sheng2024hybridflow}, which is publicly available\footnote{\url{https://github.com/RLHFlow/Online-DPO-R1/blob/main/ppo_training/verl_example.sh}}.

\textbf{Baseline: improving the self-correction ability.} We consider several baseline methods in the self-correction literature, including training-free approaches and fine-tuning. For training-free methods, we evaluate intrinsic self-correction \citep{huang2023large}, where models rely solely on prompting to perform correction, and self-correction with external ground-truth rewards \citep{qu2024recursive}. The prompts used for these methods are provided in Appendix~\ref{appendix:prompt}. We also include STaR and RAFT approaches \citep{zelikman2022star, dong2023raft}, which are inspired by expert iteration in reinforcement learning \citep{anthony2017thinking}. These methods generate numerous trajectories with the base model, filter out failed attempts, and fine-tune on successfully revised responses. Following \citet{kumar2024training}, we study a variant, STaR/RAFT+, which augments the training set with a set of correct-to-correct trajectories. To ensure a fair comparison, the total number of training samples for STaR/RAFT(+) is kept the same as in our self-rewarding IFT stage. 

\textbf{Baseline: improving the single-turn reasoning ability.} In addition, we also consider several baselines that improve the models' single-turn reasoning ability without self-correction. These methods include the STaR/RAFT \citep{zelikman2022star, dong2023raft}, iterative DPO \citep{xiong2023iterative} with the correctness score to rank data, and PPO with the correctness score. In particular, we adopt the iterative algorithms in the implementations of the STaR/RAFT and DPO because we observe that they achieve much better performance to serve as competitive baselines. We start from Qwen-2.5-Math-7B and train with only self-generated data for a fair comparison. We remark that the Qwen-2.5-Math-7B has been trained on many instruction-following data in the pre-training stage and the recent open-source projects also show that it can be used as the starting checkpoint without distillation from larger LLMs or human instructions \citep{zeng2025simplerl, zhang2025dpor1}.

\subsection{Main Results} \label{sec:qwen_main_result}
We report the main results in Table~\ref{tab:scaled_result}. Note that there can be an error of 0.1 due to rounding.

\paragraph{Intrinsic self-correction with prompting fails in general.} We first observe that intrinsic self-correction without explicit reward signals typically reduces final test accuracy. Upon analyzing the outputs, we find that models tend to modify their initial responses responses regardless of its correctness, as they lack a mechanism to determine when to refine their answers versus when to terminate the correction process. Moreover, even when given ground-truth rewards, base models with prompting alone achieve only marginal improvement in incorrect-to-correct transitions $\Delta^{i \to c}(t_1, t_2)$. For example, on MATH-500 benchmark, prompting with gold reward only leads to $\Delta^{i \to c}(t_1, t_2)=1.4\%$.

We also notice that the STaR/RAFT method, which fine-tunes models on revised incorrect attempts, fails to significantly improve performance. It increases $\Delta^{i \to c}(t_1, t_2)$ (incorrect-to-correct transitions) on MATH500 from $1.4\%$ to $5.0\%$, but still suffers from a $\Delta^{c \to i}(t_1, t_2)$ (correct-to-incorrect transitions) of $6.2\%$. Additionally, the STaR/RAFT+ variant, which includes correct-to-correct trajectories, becomes more conservative in modifying the initial attempt. While this reduces incorrect corrections ($\Delta^{c \to i}(t_1, t_2)$), it also lower $\Delta^{i \to c}(t_1, t_2)$, ultimately degrading test accuracy. These findings align with prior studies, and highlight the limitations of intrinsic self-correction, even with training \citep{huang2023large, kumar2024training}.

\begin{table*}[htp]
    \centering \footnotesize
        \caption{The results of reward modeling accuracy ($\%$). We report the accuracy of self-rewarding signals for the three benchmarks in two separate classes. For instance, MATH-500 C is the accuracy of recognizing a correct trajectory, while MATH-500 W is the accuracy of recognizing a wrong trajectory. The model highlighted by $(*)$ is selected as the final model.} \vspace{10pt} \label{tab:rm_scaled_result}
    \begin{tabular}{c|cccccc}
    \toprule
\textbf{Method}  & {MATH-500 C} & {MATH-500 W} &OlympiadBench C & {OlympiadBench W} & {Minerva Math C} & {Minerva Math W} \\ 
    \midrule 
    Self-rewarding IFT & 93.0 & 47.7 & 89.6 & 45.9 & 91.7 & 36.1 \\
   \midrule 
    PPO Step 100 & 97.5 & 56.4 & 98.1 & 33.5 & 87.4 & 29.7 \\
      PPO Step 220 $(\star)$ & 98.6 & 47.6 & 97.8 & 39.3 & 94.2 & 32.4 \\
      \midrule 
      DPO Iter 2 & 91.3 & 56.2 & 81.9 & 51.8 & 86.7 & 36.2\\
      DPO Iter 5 $(\star)$ & 92.0 & 50.6 & 88.2 & 44.5 & 92.4 & 37.4 \\
    \bottomrule
    \end{tabular}
    \end{table*}

\paragraph{Self-rewarding reasoning models significantly outperform existing baselines of self-correction.} Across all tasks, self-rewarding reasoning models consistently improve final accuracy with higher $\Delta(t_1, t_2)$ compared to baseline methods. We notice that fine-tuning on the synthetic trajectories with self-correction behavior yields models with much higher $\Delta^{i\to c}(t_1,t_2)$, suggesting that the models are more good at correcting the error in the self-generated responses. Distint from the STaR/RAFT, models trained with self-rewarding IFT also exhibit significantly lower $\Delta^{c\to i}(t_1, t_2)$, indicating they are better at recognizing when to stop due to the additional self-rewarding signals. For instance, on MATH500, 
\begin{itemize}
    \item self-rewarding IFT achieves $\Delta^{i\to c}=5.0\%$ (vs. $1.4\%$ for intrinsic self-correction);
    \item self-rewarding IFT achieves $\Delta^{c\to i}=0.4\%$ (vs. $15.4\%$ for intrinsic self-correction and $3.8\%$ for STaR/RAFT+);
\end{itemize}
Since STaR/RAFT(+) and self-rewarding IFT use the same data synthesis approach (rejection sampling) but under different self-correction frameworks, these results highlight the advantage of our self-rewarding reasoning framework.

\paragraph{Self-rewarding reasoning models improve the final accuracy compared to the single-turn baselines.} We also compare the self-rewarding reasoning models with RL training against their single-turn counterparts. For both the PPO and DPO, the self-rewarding reasoning models achieve higher final test accuracy due to the additional correction step. For instance, the self-rewarding IFT + PPO yields a model with $43.4\%$ final accuracy on OlympiadBench, and $38.4\%$ on Minerva Math, compared to the $39.5\%$ and $33.1\%$ of the single-turn counterpart. Similarly, with the DPO, the self-rewarding reasoning models achieve a $78.6\%$ on MATH500, a $40.1\%$ on OlympiadBench, and $34.6\%$ on Minerva Math, while the single-turn DPO model admits $76.8\%, 39.0\%, 31.6\%$, respectively.

However, self-rewarding models use more tokens at inference due to the additional correction step. For a fair comparison, we will also study the behavior of self-rewarding correction under scaled test-time compute budgets in Section~\ref{sec:exp}.  

\paragraph{Deep RL algorithm outperforms the direct alignment algorithms.} We observe that PPO outperforms iterative DPO by a large margin. For example, the PPO-trained model achieves a $43.4\%$ final accuracy on OlympiadBench, compared to the $40.1\%$ of the DPO method. This suggests that when absolute reward signals are available, enforcing a preference structure (Bradley-Terry model) is unnecessary and may degrade performance. Another possible reason is the limited data utilization in DPO. We notice that, with our setup, we can collect comparison pairs for only 40\% to 60\% prompts. For the remaining prompts, models either generate no correct trajectories or all trajectories are correct. As a result, DPO utilizes less training data than PPO, which may contribute to its lower accuracy.

\begin{figure*}[htp]
    \centering
    \includegraphics[width=0.32\textwidth]{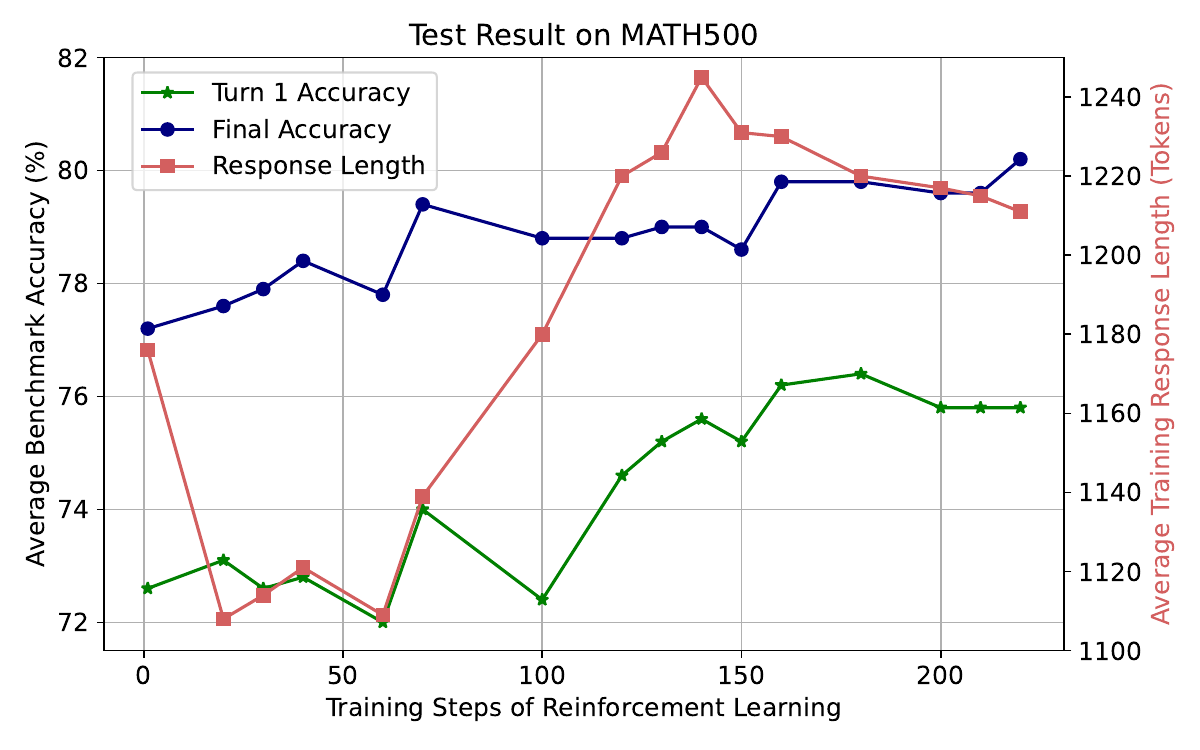}  
    \includegraphics[width=0.32\textwidth]{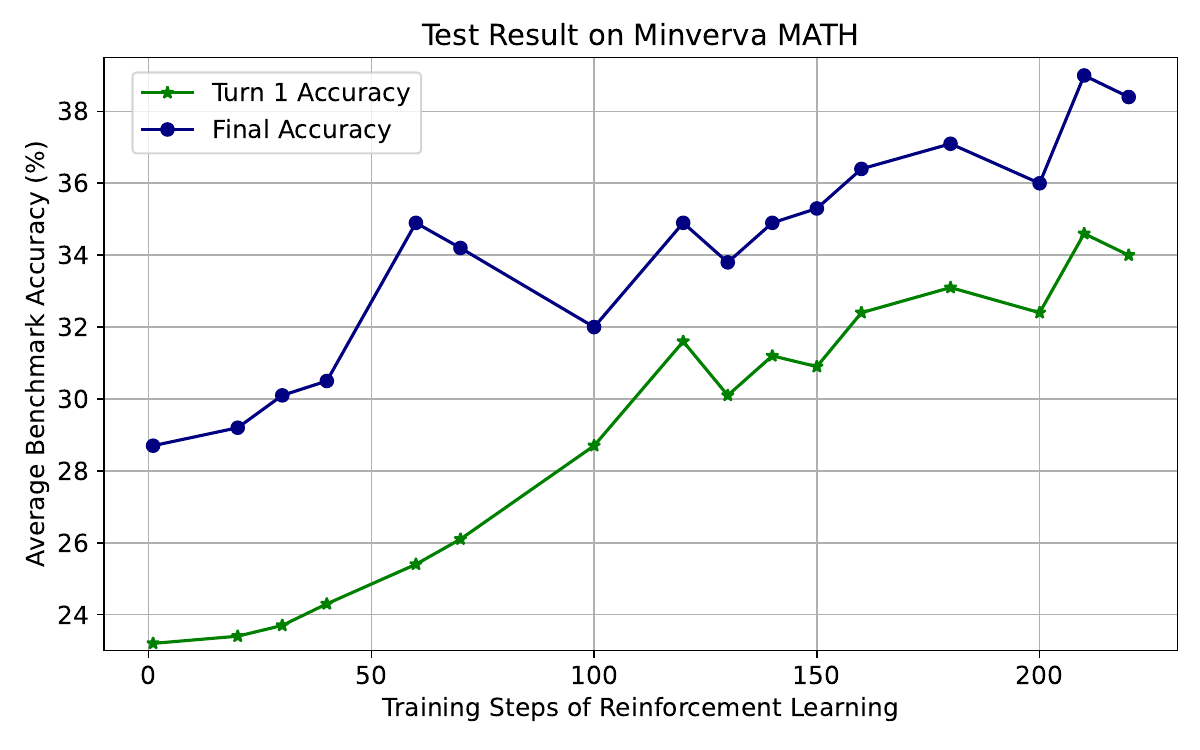} 
    \includegraphics[width=0.32\textwidth]{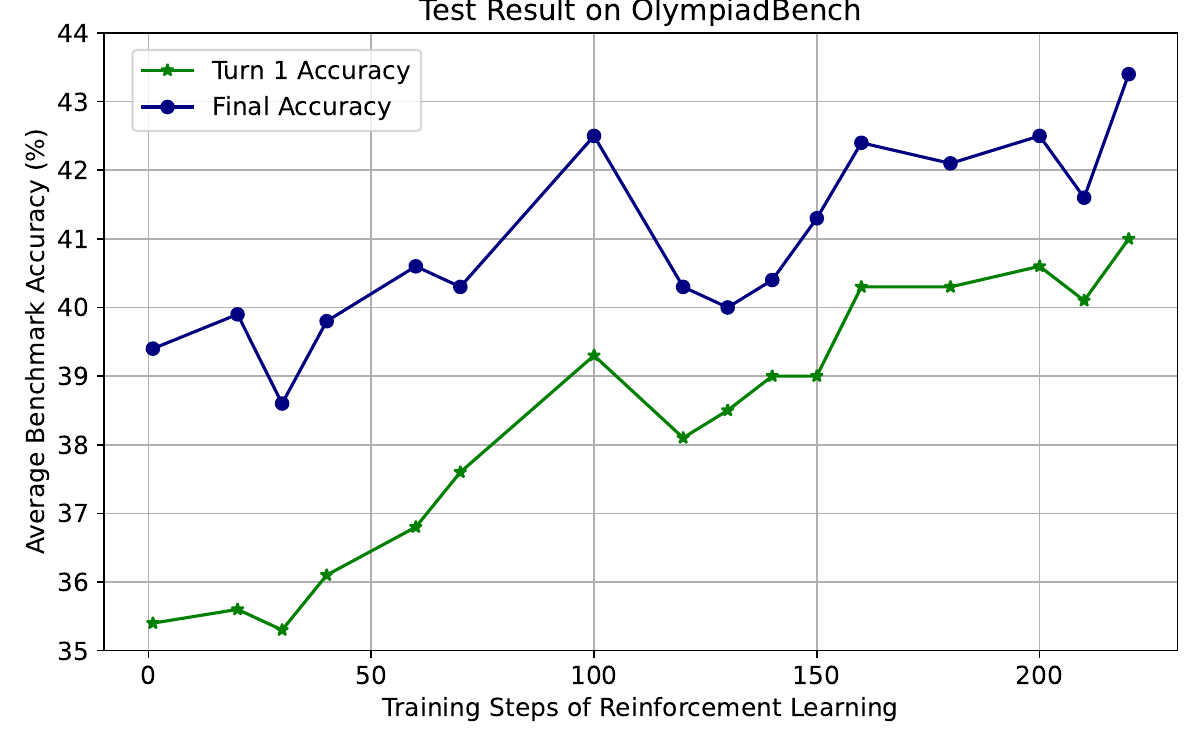}  
    \caption{The learning dynamic of the PPO training, initialized from the self-rewarding IFT model. We also plot the average generation length during the training in the first figure.}
        \label{fig:ppo} 
\end{figure*}

\paragraph{Reward model (RM) accuracy.} Since our self-rewarding framework unifies the generator and reward model, we evaluate the accuracy of our models as a reward model. We observe that the Qwen2.5-Math-7B-base can fail to strictly follow the format by omitting the self-evaluation step or not generating the evaluation result in the pre-determined format possibly because the model is not instruction-following fine-tuned. However, this happens in less then 10\% of the cases so we focus on the samples with the evaluation step and also further involve human supervision to summarize the statistics. We report the result in Table~\ref{tab:rm_scaled_result}. We observe that the self-rewarding IFT model is much more good at recognizing the correct trajectories, as the accuracy is generally higher than $90\%$, even though we balance the two types of trajectories in the training set. This directly leads to the small $\Delta^{c\to i}(t_1,t_2)$ we observe in the main table.

We also notice that the RL training (both PPO and DPO) does not consistently improve the reward modeling accuracy. Analysis of PPO checkpoints (initial model, Step 100 and Step 220) clearly shows a trade-off between correct and incorrect classification accuracy. The PPO training explores different trade-off between them, with the goal of maximizing the final accuracy. Similar observation also applies to the DPO training. Moreover, the best model of PPO training tends to prioritize recognizing correct trajectories, at the cost of lower accuracy in identifying incorrect responses, which aligns with the lower $\Delta^{c\to i}(t_1,t_2)$ and also lower $\Delta^{i\to c}(t_1,t_2)$. This may be because correcting an incorrect answer is generally more challenging than maintaining a correct initial response. We defer a more detailed study of the impact of data composition on reward modeling accuracy to the next section.

\paragraph{Learning dynamic of the RL stage.} While the RL training improves final accuracy, the final test accuracy is determined by both the turn-1 accuracy and $\Delta(t_1, t_2)$. In particular, we notice that the final accuracy gains come primarily from the higher turn-1 accuracy, as the models after the RL training usually admit a much higher turn-1 accuracy, but also a lower $\Delta^{i\to c}(t_1,t_2)$. To understand the learning dynamic of the RL training, we plot test accuracy on three benchmarks in terms of the RL training steps in Figure~\ref{fig:ppo}. We observe that in the early stage of the RL training, both the turn-1 accuracy and the final accuracy increase, and their gap $\Delta(t_1, t_2)$ is also increased or maintained as a stable level. This indicates that the models learn to use their knowledge better in the first round and improve or maintain a comparable level of correction ability. Around training step $100$, however, the increase of the final accuracy is mainly from the higher turn-1 accuracy and their gap is narrowed, indicating less reliance on self-correction.

We also plot the average generation length in the first figure. Initially, the length decreases because the Qwen2.5-Math-7B-base model tends to generate many python codes, resulting in lengthy responses. We observe that the code usually takes many tokens and can lead to incomplete reasoning path and it is discouraged by the reward signal. This observation is consistent with \citet{zeng2025simplerl}. Then, the length increases in the next stage, indicating that the reflection and self-correction abilities are also encouraged by the RL training. Finally, the length decreases again, along with a higher turn-1 accuracy and a smaller $\Delta(t_1, t_2)$, indicating that the models learn to provide a correct answer in their first attempt and also, the self-correction pattern is discouraged. This is also supported by the reward model accuracy, where the RL-trained models tend to be more conservative and evaluate the attempt as correct.

\begin{table*}[h]
    \centering \footnotesize
        \caption{Main results of different methods on the test sets of MATH (first two groups of results) and GSM8K (last two groups of results). Models are evaluated with temperature 1.0, and results are averaged over three random seeds. Additional results using a temperature of 0.7 are included in the appendix due to space constraints. } \vspace{10pt} \label{tab:main_result}
    \begin{tabular}{cc|ccccc}
    \toprule
    \textbf{Base Model} & \textbf{Method}  & \textbf{Turn 1} & \textcolor{red}{\textbf{Final Accuracy}}& \textcolor{red}{$\Delta(t_1,t_2)$} & $\Delta^{i \to c}(t_1,t_2)$ & $\Delta^{c \to i}(t_1,t_2)$ \\ 
    \midrule 
    Llama-3-8B-it & Prompt with Gold RM & 20.7 & 30.3 & 9.6 & 9.6 & 0\\
    Llama-3-8B-it & Prompt with External ORM & 20.7 & 26.2 & 5.5 & 8.8 & 3.3 \\
    Llama-3-8B-it & Intrinsic self-correction & 20.7 & 22.0 & 1.3 & 8.8 & 7.5\\
    Llama-3-8B-it & STaR/RAFT for self-correction & 22.3 & 26.1 & 3.7 & 11.4 & 7.7\\
    Llama-3-8B-it & STaR/RAFT+ for self-correctio & 22.7 & 27.1 & 4.4 & 11.7 & 7.3\\
    \rowcolor[rgb]{ .867, .922, .969} Llama-3-8B-it & Self-rewarding IFT & 22.6 & 27.9 & 5.3 & 8.8 & 3.5 \\
    \rowcolor[rgb]{ .867, .922, .969} Llama-3-8B-it & Self-rewarding IFT + Gold RM & 22.6 & 33.9 & 11.3 & 11.3 & 0 \\
    \midrule
    Llama-3-SFT & Prompt with Gold RM & 36.2 & 45.0 & 8.8 & 8.8 & 0\\
    Llama-3-SFT & Prompt with External ORM & 36.2 & 39.2 & 3.0 & 7.5 & 4.5 \\
    Llama-3-SFT & Intrinsic self-correction & 36.2 & 35.3 & -0.9 & 8.5 & 9.4\\
    Llama-3-SFT & STaR/RAFT for self-correctio & 38.5 & 36.7 & -1.8 & 10.5 & 12.3\\
    Llama-3-SFT & STaR/RAFT+ for self-correctio & 37.9 & 38.8 & 0.9 & 9.4 & 8.5\\
    \rowcolor[rgb]{ .867, .922, .969} Llama-3-SFT & Self-rewarding IFT & 37.1 & 40.3 & 3.2 & 7.2 & 4.0 \\ 
    \rowcolor[rgb]{ .867, .922, .969} Llama-3-SFT & rewarding IFT +  Gold RM & 37.1 & 46.8 & 9.7 & 9.7 & 0 \\
    \midrule
        Llama-3-8B-it & Prompt with Gold RM & 64.0 & 72.1 & 8.1 & 8.1 & 0\\
    Llama-3-8B-it & Prompt with External ORM & 64.0 & 68.0 & 4.0 & 5.9 & 1.9  \\
    Llama-3-8B-it & Intrinsic self-correction & 64.0 & 48.1 & -15.9 & 7.1 & 23.0\\
    Llama-3-8B-it & STaR/RAFT for self-correctio & 76.0 & 63.1 & -12.9 & 7.9 & 20.8\\
    Llama-3-8B-it & STaR/RAFT+ for self-correctio & 75.7 & 67.0 & -8.7 & 8.6 & 17.3 \\
    \rowcolor[rgb]{ .867, .922, .969} Llama-3-8B-it & Self-rewarding IFT & 73.2 & 78.2 & 5.0 & 9.1 & 4.1\\  
            \midrule
    Llama-3-SFT & Prompt with Gold RM & 74.6 & 83.1 & 8.5 & 8.5 & 0\\
    Llama-3-SFT & Prompt with External ORM & 74.6 & 76.7 & 2.1 & 5.5 & 3.4 \\
    Llama-3-SFT & Intrinsic self-correction & 74.6 & 67.4 & -7.2 & 7.6 & 14.8 \\
    Llama-3-SFT & STaR/RAFT for self-correctio & 73.8 & 67.4 & -6.4 & 9.0 & 15.4\\
    Llama-3-SFT & STaR/RAFT+ for self-correctio & 73.9 & 73.5 & -0.4 & 8.6 & 9.0 \\
    \rowcolor[rgb]{ .867, .922, .969} Llama-3-SFT & Self-rewarding IFT  & 76.1 & 79.2 & 3.1 & 4.7 & 1.6 \\
    \bottomrule
    \end{tabular}
    \end{table*}

\section{More Experiment Results with a Two-turn Conversation Framework and Llama Models}
\label{sec:exp}

In this section, we continue to investigate the self-rewarding reasoning framework.

\subsection{Data Format: Simplified Two-turn Framework}
Previously, we combined multiple reasoning steps into a single long CoT trajectory, which aligns with common practice. However, this approach poses significant challenges for our study, as models—particularly Qwen2.5-Math-7B-base—often fail to strictly follow instructions for evaluating or revising responses based on their history. For instance, models sometimes will also generate the evaluation results using $\boxed{}$ or not to correct the responses even though the self-evaluation result is ``[VERIFY] wrong''. Additionally, models can perform multiple rounds of self-evaluation and correction, but these steps are tightly coupled and cannot be easily decoupled into separate stages.

To address these issues, we adopt a simplified two-turn conversation framework, where the user provides explicit instructions between different steps. Specifically, after receiving the mathematical problem, the model will first generate the CoT reasoning $a^1$ and self-evaluation $y$. Then, the user provide a deterministic instruction $o$ based on the self-evaluation $y$:
\begin{enumerate}
    \item Since your initial response is self-evaluated as incorrect, there might be an error in the solution above because of lack of understanding of the question. Please correct the error, if any, and rewrite the solution. Put your final answer within $\boxed{}$;
    \item Since your initial response is self-evaluated as correct, confirm it and provide no further modifications. Put your final answer within $\boxed{}$.
\end{enumerate}
Meanwhile, when collecting the data, the self-rewarding signal is determined directly by the ground-truth oracle reward with the template designed in \citet{zhang2024generative}, without additional reasoning. While this simplification may reduce reward modeling accuracy \citep{zhang2024generative}, it facilitates controlled experimentation by allowing modifications to the self-rewarding signal. Similar frameworks—without the self-rewarding component—have been explored in previous works \citep{huang2023large, kumar2024training}. See Table~\ref{fig:example_2turn} for an illustrative example.

\begin{table*}[htp]
\centering
\begin{minipage}{2.0\columnwidth}\vspace{0mm}    \centering
\begin{sectionbox}[]{An example of the self-rewarding reasoning path under the two-turn conversation framework.} 
    \centering
      \scriptsize
\begin{tabular}{p{0.97\columnwidth} c}
\textbf{User:} Three positive integers $a,$ $b,$ and $x$ form an O'Hara triple $(a,b,x)$ if $\sqrt{a}+\sqrt{b}=x.$ For example, $(1,4,3)$ is an O'Hara triple because $\sqrt{1}+\sqrt{4}=3.$ If $(36,25,x)$ is an O'Hara triple, determine the value of $x.$ \\ Please reason step by step and put your final answer within \{\}.\\ 

    \textbf{Assistant:} Here is my solution: Begin by substituting the numbers into the definition of an O'Hara triple: $x=\sqrt{36}+\sqrt{25}$. Factor out the perfect squares from each term to simplify: $x=\sqrt{9\times4}+\sqrt{25}$. Use the perfect square roots of each term and combine them: $x=3\sqrt{4}+5$. Multiply 4 with the root: $x=12+5$. Add 12 and 5: Final Answer: The final answer is $\boxed{x=17}$. I hope it is correct. \\ \textcolor{red}{Is my most recent final answer correct (Yes or No)? No.} \\

    \textbf{User:} Since your initial response is self-evaluated as incorrect, there might be an error in the solution above because of lack of understanding of the question. Please correct the error, if any, and rewrite the solution.  \\

    \textbf{Assistant:} Let's re-evaluate the solution: We are given that (36, 25, x) is an O'Hara triple, so we can write: $\sqrt{a} + \sqrt{b} = x$. Substituting a = 36 and b = 25, we get: $\sqrt{36} + \sqrt{25} = x$. Simplifying the square roots, we get:$6 + 5 = x$. Adding 6 and 5, we get: $\boxed{x = 11}$
    \end{tabular}
\end{sectionbox}
\caption{An example of the self-rewarding reasoning path under the two-turn conversation framework.}
    \label{fig:example_2turn}
\end{minipage}
\end{table*}

\subsection{Experiment Setup}

\paragraph{Base model, task, and datasets.} Qwen2.5-Math-7B-base serves as a strong and specialized base model, which is pre-trained on a large mathematical corpus. To ensure generality and a more comprehensive evaluation, we experiment with the Llama model series. Specifically, our base models include Llama-3-8B-it and Llama-3-SFT, the latter being fine-tuned on Open-MathInstruct2-1M \citep{toshniwal2024openmathinstruct2}. While both models are generally weaker than Qwen2.5-Math-7B-base, Llama-3-SFT is stronger than Llama-3-8B-it.

In this section, we evaluate the models' mathematical reasoning abilities using the MATH and GSM8K benchmarks, which are well-suited to their capacities. For MATH, we use 7.5K training problems during the self-rewarding IFT stage, supplemented by 7.5K prompts from Open-MathInstruct2 for M-DPO training, with a similar setup for GSM8K. Model selection is performed using a 1K validation set from Open-MathInstruct2. Since we formulate the task as a multi-turn chat problem, we can directly use Axolotl's training code\footnote{\url{https://github.com/axolotl-ai-cloud}}. During inference, the model generates up to 2048 tokens per round, with VLLM 0.5.4 \citep{kwon2023efficient} accelerating the process.

\paragraph{Training Setup for Llama SFT.}
For the self-rewarding IFT stage, we use a learning rate of 2e-6 with a batch size of 32 for Llama models and 64 for Llama-3-SFT training. Outcome-supervised reward models (ORMs) are trained using standard SFT recipes and datasets, as described in \citep{xiong2024implementation}. Full hyperparameter configurations will be available in our GitHub repository.

We observe that models occasionally fail to follow the instruction to perform self-rewarding corrections, though this occurs in less than 5\% of cases. In such scenarios, we terminate after the first round and use its output as the final answer. 

\subsection{Main Results with Llama Models}

\textbf{Experiments with Llama models align well with the Qwen model.} Our experiments with Llama models show similar trends to those observed with Qwen models. Specifically, intrinsic self-correction—whether with or without STaR/RAFT-like training—fails to reliably correct errors in self-generated responses. Models tend to modify their initial responses regardless of correctness, making these methods beneficial primarily for weaker models where most first attempts are incorrect (e.g., MATH task with Llama-3-8B-it). However, for stronger models that solve most problems correctly on the first attempt (e.g., GSM8K task with Llama-3-SFT), intrinsic self-correction and STaR/RAFT methods significantly reduce turn-2 accuracy. In contrast, self-rewarding IFT models consistently improve turn-1 accuracy by effectively correcting errors while preserving already correct responses. This demonstrates the generality of the proposed framework. 

To further evaluate the self-rewarding IFT model, we modify the self-rewarding signal to be the same as the oracle reward, eliminating the influence of reward signal quality and directly assessing the model's ability to correct incorrect responses. For example, the baseline Llama-3-SFT achieve a $\Delta^{i\to c}(t_1, t_2)=8.8\%$, while models fine-tuned with self-rewarding IFT exhibit a higher $\Delta^{i\to c}(t_1, t_2)=9.7$, indicating improved correction capabilities.

\subsection{Self-rewarding Reasoning Framework Improves Efficiency in Test-time Inference Compute Scaling} 
\begin{figure*}[htp]
    \centering
    \includegraphics[width=0.48\textwidth]{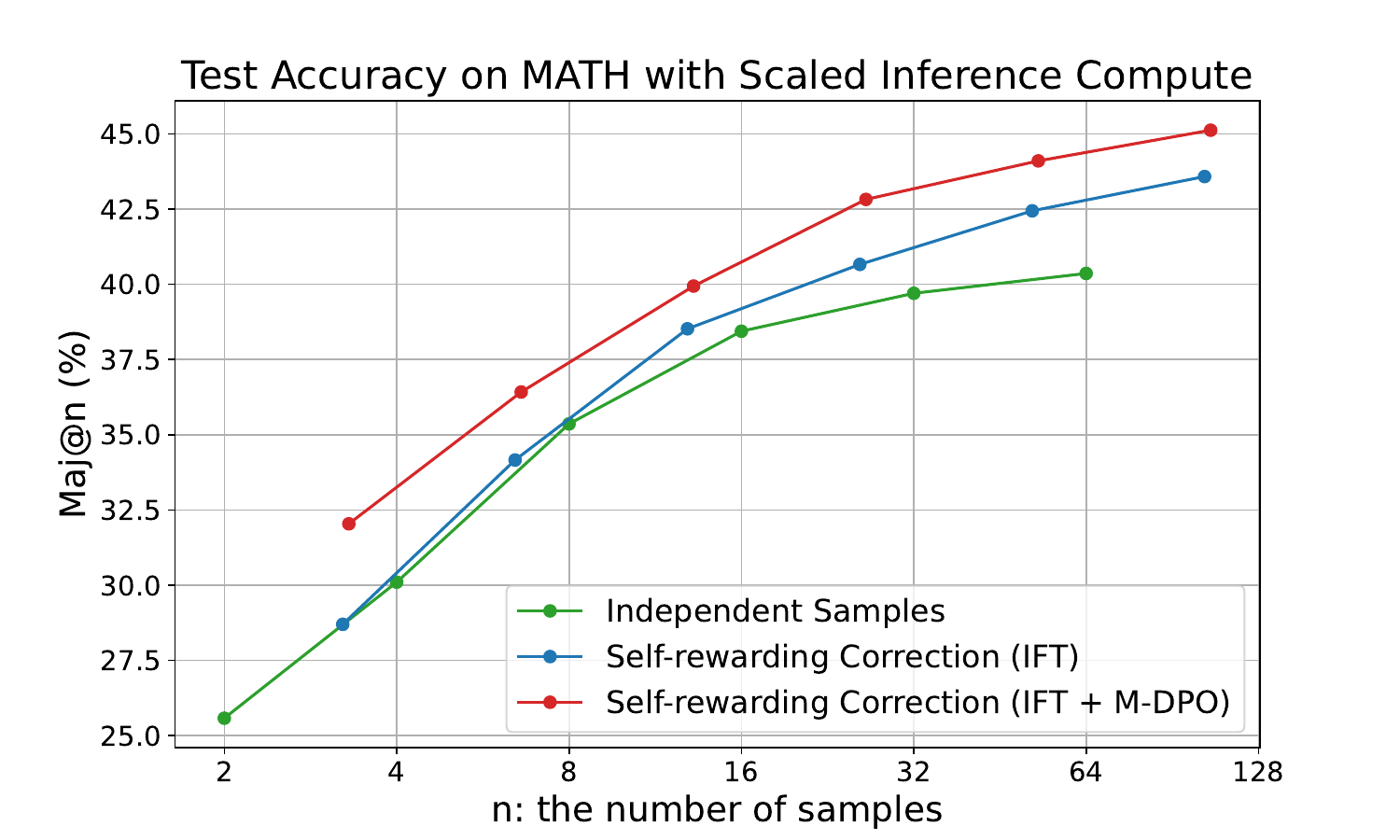}  
    \includegraphics[width=0.48\textwidth]{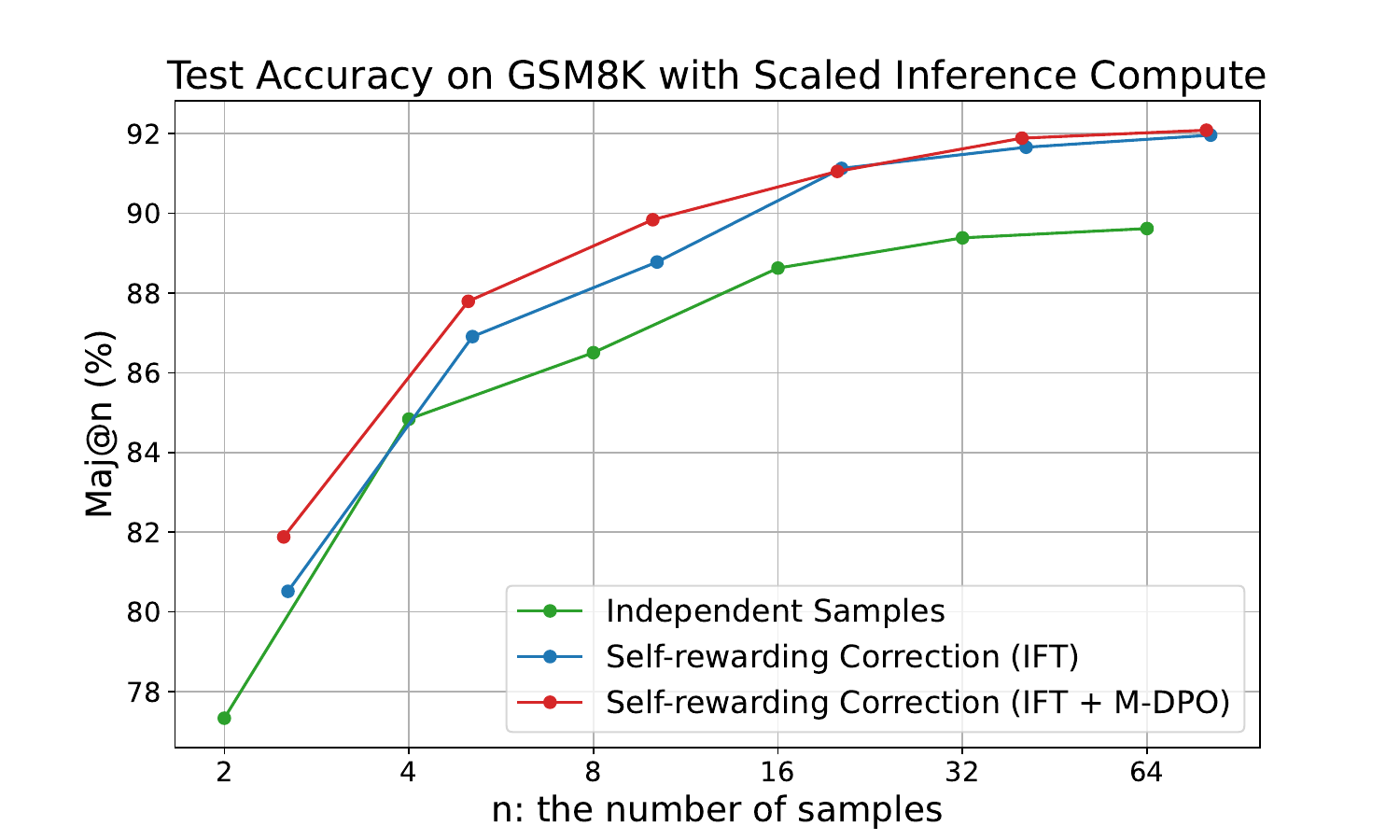}  
    \caption{The majority voting results of independent samples and self-rewarding correction with Llama-3-8B-it. For MATH, we collect 1.61 samples per trajectory on average with our IFT model, and 1.65 samples per trajectory on average with our M-DPO aligned model, and for GSM8K, we collect 1.27 samples per trajectory for the IFT model and 1.25 samples for the M-DPO aligned model.}
        \label{fig:inference_time}
\end{figure*}

Self-correction requires generating multiple LLM responses, making it crucial to compare models under equivalent inference budgets. As noted by \citet{huang2023large}, prior self-correction approaches often perform no better than self-consistency \citep{wang2022self} when constrained to an equal number of responses. To address this, we analyze self-rewarding correction under scaled test-time compute budgets by sampling $N$ reasoning paths and using majority voting for the final output. We report the results in Figure~\ref{fig:inference_time}, where the DPO-aligned model is descripted in Section~\ref{sec:llama_dpo}. For both MATH and GSM8K tasks, with a fixed inference budget, the self-rewarding correction model consistently outperforms independent sampling methods. For example, the independent sampling achieves an accuracy of 40.4\% on MATH with 64 samples, whereas the self-rewarding correction method (using IFT and M-DPO training) achieves an accuracy of 42.8\% with only 26.4 samples. 

One key factor contributing to this improved efficiency is that, unlike intrinsic self-correction or STaR/RAFT methods, our models do not necessarily generate two samples per trajectory. Instead, they terminate early when the model is confident in the correctness of its first-round response. For instance, using Llama-3-8B-it as the base model, our approach generates an average of 1.65 samples per trajectory for MATH and 1.25 samples per trajectory for GSM8K, leading to significant computational savings.

\subsection{Ablation Study on Data Distribution}

\textbf{Self-rewarding IFT models outperforms the self-correction with external ORMs.} To better understand the dynamics of the self-rewarding signal, we compare self-rewarding IFT models to an external ORM trained on the same dataset, with results reported in Table~\ref{tab:compare_external_orm}. We observe that self-rewarding IFT models achieve superior performance in both turn-2 accuracy and $\Delta(t_1, t_2)$ compared to self-correction with external ORMs. This highlights the potential of unifying the generator and reward model within a single LLM.

However, we also observe that there is a considerable gap in the reward model accuracy between the external ORM (used to evaluate Llama-3-SFT policy) and the self-rewarding RM (used to evaluate the self-rewarding IFT policy). Specifically, the self-rewarding IFT method (self-rewarding IFT policy + self-rewarding RM), achieves an accuracy of $70.0\%$ in recognizing a correct trajectory, which is slightly higher than the $66.9\%$ of the Llama-3-SFT policy + external ORM. In contrast, for the trajectories with wrong answer, the accuracy of the self-rewarding IFT model is $76.4\%$, which is much lower than the $88.4\%$ of the Llama-3-SFT policy + external ORM.

\begin{table*}[htp]
    \centering \footnotesize
        \caption{Comparison between self-rewarding IFT models and Llama-3-SFT model with external ORM on MATH benchmark. We report the accuracy of self-rewarding signals for the three benchmarks in two separate classes. For instance, MATH C is the accuracy of recognizing a correct trajectory, while MATH W is the accuracy of recognizing a wrong trajectory.} \vspace{10pt} \label{tab:compare_external_orm}
    \begin{tabular}{c|ccccccc}
    \toprule
\textbf{Method}  & \textbf{Turn 1} & \textbf{Final Accuracy} & $\Delta(t_1,t_2)$ & $\Delta^{i \to c}(t_1,t_2)$ & $\Delta^{c \to i}(t_1,t_2)$ & MATH C & MATH W \\ 
    \midrule 
    Llama-3-SFT + Gold RM & 36.2 & 45.0 & 8.8 & 8.8 & 0 & 100 & 100 \\
        Llama-3-SFT + External ORM  & 36.2 & 39.2 & 3.0 & 7.5 & 4.5 & 66.9 & 88.4 \\
    Llama-3-SFT + Self-rewarding RM  & 36.2 & 38.9 & 2.7 & 7.4 & 4.7 & 67.0 & 86.7\\
    \midrule 
  Self-rewarding IFT + Self-rewarding RM & 37.1 & 40.3 & 3.2 & 7.2 & 4.0 & 70.0 & 76.4 \\ 
   Self-rewarding IFT + Gold RM & 37.1 & 46.8 & 9.7 & 9.7 & 0 & 100 & 100\\
    \bottomrule
    \end{tabular}
    \end{table*}

To better understand this discrepancy, we use the self-rewarding RM to guide the self-correction of the Llama-3-SFT policy. Interestingly, under this setting, the reward model accuracy for Llama-3-SFT aligns more closely with that of the external ORM, suggesting the presence of an out-of-distribution (OOD) issue. Specifically, the policy shifts from Llama-3-SFT to self-rewarding IFT policy during self-rewarding IFT stage, while the reward model is trained on data generated by the original Llama-3-SFT policy. Furthermore, even when evaluating the same Llama-3-SFT policy with both the self-rewarding RM and the external ORM, we observe that self-rewarding training slightly degrades the reward model's capability, primarily due to the capacity limitations of the model. We believe that addressing the OOD issue and using a larger base model could further enhance the performance of self-rewarding models, which we leave for future work.

\paragraph{Data composition in Self-rewarding IFT influence the ORM accuracy.}  In our experiments with Qwen and Llama models, even though we use balanced training set with equal numbers of trajectories with incorrect first answers ($\cD^{\mathrm{IFT}}_1$) and correct first answers ($\cD^{\mathrm{IFT}}_3$), the reward modeling accuracy in two classes are unbalanced. Moreover, while the Qwen model is better at recognizing the correct trajectory (see Table~\ref{tab:rm_scaled_result}), the Llama model is better at recognizing the wrong trajectory (see Table~\ref{tab:compare_external_orm}). To analyze this further, we conduct an ablation study on dataset composition, testing three variations using the Llama-3-8B-it model:
\begin{enumerate}
    \item Balanced training set: equal numbers of trajectories with incorrect first answers ($\cD^{\mathrm{IFT}}_1$) and correct first answers ($\cD^{\mathrm{IFT}}_3$);
    \item More incorrect trajectories: $|\cD^{\mathrm{IFT}}_1| > |\cD^{\mathrm{IFT}}_3|$;
    \item More correct trajectories: $|\cD^{\mathrm{IFT}}_3| > |\cD^{\mathrm{IFT}}_1|$.
\end{enumerate}
We also investigate the impact of the additional correct-to-correct trajectories. Our findings are reported in  Table~\ref{tab:ablation_data_llama3_sft}. 

The results indicate that, for a fixed number of incorrect trajectories, increasing the proportion of correct trajectories (e.g., transitioning from a dataset with more incorrect trajectories to a balanced dataset) biases the ORM toward predicting answers as correct. This results in higher accuracy for recognizing correct trajectories but lower accuracy for identifying incorrect ones. Specifically, from a balanced training set to the training set with more correct trajectories, the accuracy changes from $(72.1\%, 75.3\%)$ to $(63.6\%, 82.4\%)$. This highlights a trade-off between these class-dependent accuracies as the changes in the reward model’s accuracy directly influence the transitions between correct and incorrect answers.

\begin{table*}[h]
    \centering \scriptsize
        \caption{Ablation study on the training sets of self-rewarding IFT with the base model Llama-3-8B-SFT. For the balanced training set, we use 125K trajectories with incorrect first answers ($\cD^{\mathrm{IFT}}_1$) and 125K with correct first answers ($\cD^{\mathrm{IFT}}_3$).  For sets with more incorrect trajectories, $|\cD^{\mathrm{IFT}}_1| = 125K$ and $|\cD^{\mathrm{IFT}}_3| = 60K$. Finally, for the training set with more correct trajectories, we have $|\cD^{\mathrm{IFT}}_1| = 125K$ and $|\cD^{\mathrm{IFT}}_3| = 180K$. Models trained with more incorrect trajectories (marked by $(\star)$) are used as final model and the dataset is also used to train the external ORM. ``+ c2c 60K'' indicates an additional 60K correct-to-correct trajectories and ``+Gold RM'' replaces self-rewarding signals with ground-truth labels during inference.} \vspace{5pt} \label{tab:ablation_data_llama3_sft}
    \begin{tabular}{c|ccccccccc}
    \toprule
     \textbf{Method}  & \textbf{Turn 1} & \textbf{Final Accuracy} & $\Delta(t_1,t_2)$ & $\Delta^{i \to c}(t_1,t_2)$ & $\Delta^{c \to i}(t_1,t_2)$ & $p^{c \to i}(t_1, t_2)$ & \textbf{RM Accuracy} \\ 
    \midrule 
    Llama-3-SFT + Gold RM & 36.2 & 45.0 & 8.8 & 8.8 & 0 & - & (100, 100)\\
        Llama-3-SFT + External ORM $(\star)$ & 36.2 & 39.2 & 3.0 & 7.5 & 4.5 & 37.6 & (66.9, 88.4) \\
    Llama-3-SFT + Self-rewarding RM $(\star)$  & 36.2 & 38.9 & 2.7 & 7.4 & 4.7 & 39.4 & (67.0, 86.7)\\
    \midrule 
  Self-rewarding IFT +  Balanced $(\star)$ & 37.4 & 40.1 & 2.7 & 7.4 & 4.7 & 45.0 & (72.1, 75.3)\\ 
     \qquad + c2c 60K & 37.1 & 40.3 & 3.2 & 7.2 & 4.0 & 36.1 & (70.0, 76.4) \\ 
   \qquad + Gold RM & 37.1 & 46.8 & 9.7 & 9.7 & 0 & - & (100, 100)\\
                  \midrule 
    Self-rewarding IFT + More Incorrect  & 38.1 & 40.3 & 2.2 & 8.0 & 5.8 & 41.7 & (63.6, 82.4)\\
      \qquad + c2c 60K  &  37.7 & 40.8 & 3.1 & 8.0 & 4.7& 33.0 &  (61.5, 84.3)\\ 
      \qquad + Gold RM & 37.7 & 46.9 & 9.2 & 9.2 & 0 & - & (100, 100) \\
     \midrule
          Self-rewarding IFT + More Correct  & 37.8 & 40.5 & 2.7 & 7.4 & 4.7 & 45.2 & (72.6, 75.1) \\ 
      \qquad + c2c 60K & 37.9 & 40.8 & 2.9 & 6.6 & 3.7 & 35.2 & (72.1, 76.2) \\ 
            \qquad + Gold RM & 37.9 & 47.5 & 9.6 & 9.6 & 0 & - & (100, 100) \\ 
    \bottomrule
    \end{tabular}
    \end{table*}
    
Comparing the results with and without the additional correct-to-correct trajectories, we observe that the additional correct-to-correct trajectories mainly contribute to a lower $p^{c \to i}(t_1, t_2)$, which is the probability of modifying a correct answer to incorrect when facing a misleading reward. This indicates that the models become more conservative when modifying initial responses. This behavior is reasonable, as correcting an incorrect answer is generally more challenging than maintaining a correct initial response.

\paragraph{The impact of distillation.} Although we focus on on-policy training, meaning that we train the models on the self-generated data only, we also try to use the Llama-3.1-70B-it to generate $a^2$ in the self-rewarding IFT of Llama-3-8B-it, with results shown in Table~\ref{tab:ablation_dpo}. We observe that teacher model data can significantly boosts turn-1 accuracy, leading to higher turn-2 accuracy. However, stronger $a^2$ does not lead to a higher $\Delta^{i \to c}(t_1, t_2)$, meaning that the models' abilities to self-correct are similar. Off-policy training also causes a substantial distribution shift in $a^1$, reducing reward model accuracy ($36.7\%$ v.s. $63.6\%$). Thus, distillation is better suited for improving turn-1 accuracy, while self-generated data is more effective for building self-rewarding reasoning models when a teacher model is available.

\subsection{Additional Rule Designs in RL Training}
\label{sec:llama_dpo}

We also conduct preliminary experiments on reward assignment strategies for PPO training and data ranking strategies for DPO training to analyze their impact on model performance.

\paragraph{The impact of ranking strategies in multi-turn DPO training\footnote{In implementation, the only difference of the multi-turn DPO and the regular DPO is that we mask out the external instruction. See \citet{xiong2024building} for the detailed derivation.}.}  For a fixed $(x, a^1)$, we experiment with the following ranking strategies:
\begin{itemize}[itemsep=2pt]
    \item $\cD^{\mathrm{M-DPO}}_1$: (wrong $a^1$, $y = \text{No}$, correct $a^2$) $\succ$ (wrong $a^1$, $y = \text{No}$, wrong $a^2$);
    \item $\cD^{\mathrm{M-DPO}}_2$: (correct $a^1$, $y = \text{No}$, correct $a^2$) $\succ$ (correct $a^1$, $y = \text{No}$, wrong $a^2$);
    \item $\cD^{\mathrm{M-DPO}}_3$: (wrong $a^1$, $y = \text{No}$, correct $a^2$) $\succ$ (wrong $a^1$, $y = \text{Yes}$);
    \item $\cD^{\mathrm{M-DPO}}_3$: (correct $a^1$, $y = \text{Yes}$) $\succ$ (correct $a^1$, $y = \text{No}$, wrong $a^2$). \vspace{-5pt}
\end{itemize}
We group the last two types of data into $\cD^{\mathrm{M-DPO}}_3$ because they involve the reward signal comparison. We exclude comparisons like (wrong $a^1$, $y = \text{No}$, wrong $a^2$) and (wrong $a^1$, $y = \text{Yes}$) as their comparison can be ambiguous. For simplicity, we only train the model for one iteration. We report the results in Table~\ref{tab:ablation_dpo}. 

Across various base models and tasks, we observe that M-DPO training with $\cD_2^{\mathrm{M-DPO}}$ effectively reduces the $p^{c \to i}(t_1, t_2)$, making models more conservative when incorrectly classifying a correct initial answer as incorrect. Consequently, models fine-tuned with M-DPO exhibit significantly lower $\Delta^{c \to i}(t_1,t_2)$, e.g., on MATH, it drops from $3.5\%$ to $2.8\%$, and on GSM8K, from $4.1\%$ to $2.5\%$. Accordingly, the M-DPO method further enhances self-rewarding reasoning language models, improving the turn-2 accuracy and $\Delta(t_1, t_2)$. Interestingly, even though the generation of $a^1$ is not explicitly involved during training, the correction ability in turn 2 naturally transfers, leading to higher turn-1 accuracy.

\begin{table*}[htp]
    \centering \scriptsize
        \caption{Ablation study on the impart of training sets of M-DPO and distillation, with Llama-3-8B-it as the base model. } \vspace{5pt} \label{tab:ablation_dpo}
    \begin{tabular}{l|ccccccccc}
    \toprule
     \textbf{Method}  & \textbf{Turn 1} & \textbf{Final Accuracy} & $\Delta(t_1,t_2)$ & $\Delta^{i \to c}(t_1,t_2)$ & $\Delta^{c \to i}(t_1,t_2)$ & $p^{c \to i}(t_1, t_2)$ & \textbf{Accuracy} \\ 
    \midrule 
   Self-rewarding IFT (MATH) & 22.6 & 27.9 & 5.3 & 8.8 & 3.5 & 43.9 & (63.6, 76.1)\\ 
    \qquad + M-DPO with $\cD_{1}$ & 24.9 & 29.1 & 4.2 & 9.3 & 5.1 & 50.3 & (59.2, 77.1)\\
    \qquad + M-DPO with $\cD_{2}$ & 24.2 & 27.8 & 3.6 & 5.5 & 1.9 & 31.3 & (74.7, 65.8)\\
        \qquad + M-DPO with $\cD_{1,2}$ & 23.9 & 28.6 & 4.7 & 6.5 & 1.8 & 27.5 & (73.4, 68.6)\\
    \qquad + M-DPO with $\cD_{1,2,3}$ (well-tuned) & 23.3 & 29.9 & 6.6 & 9.4 & 2.8 & 34.2 & (61.6, 81.4)\\
     Self-rewarding IFT + Distillation (MATH) & 28.3 & 30.5 & 2.2 & 8.0 & 5.8 & 37.5 & (36.7, 76.7)\\ 
    \midrule 
   Self-rewarding IFT (GSM8K) & 73.2 & 78.2 & 5.0 & 9.1 & 4.1 & 26.3 & (79.3, 74.0)\\ 
   \qquad + M-DPO with $\cD_{1}$ & 75.3 & 79.1 & 3.8 & 8.1 & 4.3 & 31.1 & (82.1, 70.1)\\
   \qquad + M-DPO with $\cD_{2}$ & 74.6 & 79.9 & 5.3 & 7.1 & 1.8 & 12.5 & (80.3, 70.4)\\
        \qquad + M-DPO with $\cD_{1,2}$ & 74.6 & 81.0 & 6.4 & 8.9 & 2.5 & 18.8 & (82.3, 69.6)\\
    \qquad + M-DPO with $\cD_{1,2,3}$ & 74.9 & 80.7 & 5.8 & 8.6 & 2.8 & 15.8 & (76.7, 67.1) \\
    \bottomrule
    \end{tabular}
    \end{table*}

However, we also notice that when exceeding a certain region, the excessively low $p^{c \to i}(t_1, t_2)$ can make models overly conservative, ultimately reducing the correction rate $\Delta^{i \to c}(t_1,t_2)$. This is validated in experiments using only $\cD_2^{\mathrm{M-DPO}}$, where $\Delta^{i \to c}(t_1,t_2)$ decreases from $8.8\%$ to $5.6\%$ on MATH. Conversely, training with $\cD_1^{\mathrm{M-DPO}}$ encourages models to modify their initial responses, reflected in a higher $p^{c \to i}(t_1, t_2)$, and slightly enhances the ability of correction. We notice that while on GSM8K, the model trained with $\cD_1^{\mathrm{M-DPO}}$ admits a lower $\Delta^{i \to c}(t_1,t_2)$, it mainly results from the lower RM accuracy and the higher turn-1 accuracy. If we consider the ratio of corrected trajectories, self-rewarding IFT achieves $45.9\%$, while the M-DPO-aligned model slightly outperforms it at $46.4\%$. Moreover, the combination of $\cD_1^{\mathrm{M-DPO}}$ and $\cD_2^{\mathrm{M-DPO}}$ often yields near-optimal results, striking a balance by making models more aware of when to change their initial responses.

\paragraph{DPO training cannot consistently improve the reward model accuracy.} During the experiments, we observe that M-DPO training also shifts the generation distribution of $a^1$, impacting reward model accuracy unpredictably. For example, on MATH, using only $\cD_1^{\mathrm{M-DPO}}$ reduces recognition of correct answers, while combining $\cD_1^{\mathrm{M-DPO}}$ with $\cD_2^{\mathrm{M-DPO}}$ improves recognition of correct answers but decreases accuracy for other classes by $10\%$. 

Even though we include the comparison pairs in $\cD_3^{\mathrm{M-DPO}}$, with our best efforts in tuning the dada combination in this dataset, we still suffer from the performance drop in correct answer recognition. Moreover, with simple balanced $\cD_3^{\mathrm{M-DPO}}$, such as in GSM8K, the reward model accuracy in two classes gets worse. In either case, the reward model accuracy is not consistently improved. We suspect that this is because of the mismatch of the DPO implicit reward ($\log \frac{\pi}{\pi_{\reff}}$) and the sampling probability $\log \pi$. Exploring algorithms like SimPO \citep{meng2024simpo}, which directly optimize $\log \pi$, is a promising direction for future work. Similarly, for PPO training, one may also need to adopt a multi-turn design, while we only impose KL regularization on partial responses and allow the model to adjust the self-rewarding stage more easily.

\paragraph{Additional rule designs in PPO training.} We also investigate different reward signal designs in PPO training, aiming to enhance self-correction, particularly in the later training stages. Specifically, we experiment with the following two approaches: 
\begin{enumerate}
    \item If the first attempt is incorrect and the final answer is correct, assign a reward of 1.5. Otherwise, assign 1.0 for a correct final answer and 0.0 for an incorrect one.
    \item We divide the learning in two stages. In the first stage, we train with only correctness-based rewards. Then we initialize the model from stage 1, and apply the modified reward assignment from the first plan.
\end{enumerate}
We observe that the models easily hack the first reward design, where they deliberately predict a wrong answer in the first attempt and then correct it in the second round. For instance, after 150 steps of PPO training, test accuracy on MATH500 is $18.6\%$ on first attempts but $77.6\%$ on final answers, demonstrating clear exploitation of the reward shortcut. Meanwhile, the models also struggle with the two-staged learning in plan 2, and we do not observe test accuracy improvement. 

While naive reward modifications fail, we expect that more sophisticated multi-turn RL strategies such as SCoRe \citep{kumar2024training} could further improve RL training. However, implementing multi-turn deep RL training and decoupling self-rewarding reasoning steps remains challenging in open-source frameworks, which we leave for future exploration.

\section{Conclusion and Future Research Direction}

In this work, we introduce the self-rewarding reasoning framework for LLMs, demonstrating its effectiveness in enhancing self-correction capabilities and computational efficiency. By integrating self-rewarding IFT and reinforcement learning, our approach enables LLMs to detect errors in their reasoning paths and refine their responses based on historical attempts and self-rewarding signals. Experimental results show that this framework significantly outperforms intrinsic self-correction, highlighting its potential as a robust and efficient solution for enhancing LLM reasoning.

There are still many interesting directions to improve the performance of the self-rewarding reasoning framework. First, current models show lower reward model accuracy compared to external ORMs, likely due to distribution shifts and model capacity limitations. Techniques like model merging \citep{rame2024warm, lin2023speciality} may address these issues. While we can boost the self-rewarding IFT stage by both the PPO and iterative DPO algorithms with the correctness score, our study indicates that in the late stage of RL training, the self-correction ability is not well enhanced. While our preliminary attempts on modifying the rule-based reward signals fail, we expect that incorporating multi-turn RL methods \citep{kumar2024training, shani2024multi} with the adjusted rule designs could further enhance model performance. Finally, extending beyond turn-based self-rewarding correction to step-wise correction (similar to outcome-supervised and process-supervised rewards) may offer more advantages and lead to a more scalable and dynamic approach to reasoning.

\section*{Impact Statement}

This paper presents work whose goal is to advance the field of mathematical reasoning for large language model. We proposed a self-rewarding framework to integrate the reward model and generator into a single LLM. The proposed framework can help us to build stronger LLM models in the face of complex decision making problems, thus making LLMs more helpful and contributing to the welfare of society.

\nocite{langley00}

\bibliography{example_paper}
\bibliographystyle{icml2025}

\newpage
\appendix
\onecolumn

\section{Extended Related Works} \label{sec:extended_related_work}

\paragraph{LLMs for Mathematical Problem Solving.} LLMs have shown great capacity in reasoning-related mathematical problem solving tasks \citep{cobbe2021gsm8k, hendrycks2021measuring, OpenAI2023GPT4TR, team2023gemini}. To elicit the reasoning ability of LLMs Chain-of-Thought (CoT) prompting \citep{wei2022chain, zhou2022least, zhu2022solving} has been used as a standard approach, enabling step-by-step reasoning. Another line of work equips the LLMs with external tools like calculators \citep{cobbe2021training, shao2022chaining}, symbolic solvers \citep{zhang2023mathematical}, and python interpreters \citep{mishra2022lila, OpenAI2023GPT4TR}. These LLM agents with external tools achieve even further impressive reasoning ability across a wide range of reasoning tasks \citep{chen2022program, gao2023pal, gou2023tora}. While we focus on the CoT scenario, the proposed framework and algorithms also naturally apply to the tool-integrated reasoning, which we leave for future work. 

\paragraph{RLHF for Mathematical Problem Solving.} In recognition of the tremendous successes of RL in making the general-purpose chatbot, researchers have worked on adapting these methods to building strong mathematical reasoning models. These algorithms can be largely grouped into three different categories. Among them, reward-ranked fine-tuning (or rejection sampling fine-tuning) \citep{dong2023raft, yuan2023rrhf, touvron2023llama, zelikman2022star} is extensively employed for synthetic data generation, whether through on-policy (self-improving) \citep{yuan2023scaling} or off-policy (knowledge distillation) methods \citep{gou2023tora, yu2023metamath, toshniwal2024openmathinstruct, singh2023beyond, tongdart}. These methods typically generate a large amount of trajectories and use a reward model (either through final result checking or an outcome supervised reward model) to select samples for further fine-tuning. Another line of works uses the deep-RL methods such as PPO \citep{schulman2017proximal} or Reinforce variants \citep{williams1992simple}. For instance, \citet{shao2024deepseekmath} proposes the GRPO algorithms to improve the multi-turn math problem solving in the CoT format and achieves the state-of-the-art performance in its class. \citet{kumar2024training} adopts a variant of \citep{ahmadian2024back} to improve the self-correction ability of models. Finally, a line of works apply the direct preference learning algorithms to mathematical problem solving mainly because of its simplicity and computational efficiency \citep{jiao2024learning, yuan2024advancing, xie2024monte, pang2024iterative, lai2024step, chen2024step, lu2024step}. Most of these works focus on the single-turn scenario and apply the original DPO \citep{rafailov2023direct} or KTO \citep{ethayarajh2024kto} algorithms. After these, \citet{xie2024exploratory, zhong2024dpo, xiong2024building, rafailov2024r} extend the single-turn DPO to multi-turn scenario with trajectory preference. Our algorithm is a combination of the reward-ranked fine-tuning (the self-rewarding IFT stage) and direction preference learning (the M-DPO stage) and the main focus of the algorithmic design in this project is to adapt them into the self-rewarding reasoning agent framework, with the representative self-correction task.

\section{Missing Experimental Details}
\label{appendix:prompt}

\paragraph{Prompt Template.} We present the prompt template used in our experiments here, where we mainly follow the prompt design in \citet{kumar2024training} with slight modifications.

Self-rewarding prompt used in the two-turn conversation framework: Since your initial response is self-evaluated as incorrect, there might be an error in the solution above because of lack of understanding of the question. Please correct the error, if any, and rewrite the solution.

Intrinsic self-correction: There might be an error in the solution above because of lack of understanding of the question. Please correct the error, if any, and rewrite the solution.

Gold Test: Your initial response is evaluated as incorrect. There might be an error in the solution above because of lack of understanding of the question. Please correct the error, if any, and rewrite the solution.

\paragraph{Python Experiment Environment.} The python package versions and virtual machine we use can influence the evaluation result. While this does not affect the overall trend, we specify the key package versions we use here. For IFT and M-DPO training for the Llama models, we use transformers 4.44.1 and torch 2.1.2. For IFT, we use the open-source axolotl project with version 0.4.1 and for M-DPO, we use the code base from the original M-DPO paper \citep{xiong2024building}. The setup for the Qwen models is similar, except for an updated axoltol 0.6.0 (to use the new models). For PPO training, we use the veRL v0.1. We use sympy 1.2, antlr4-python3-runtime 4.11.0, following \citet{gou2023tora} for the result checking. We use VLLM 0.5.4 to generate completions. For Llama-3-8B-it model evaluation, we also use the transformers 4.44.1, while for Llama-3-SFT-based experiments, we fix the transformers to be 4.46.1 because one of our machine was unavailable during preparing the draft of this work and we upgrade transformers to fix some bugs in a new machine.

\section{Additional Experimental Results}
\label{appendix:more_results}
In this section, we include additional ablation studies and evaluation results for a more comprehensive understanding of the self-rewarding reasoning framework and the proposed algorithms.

\begin{table}[htp]
    \centering \footnotesize
        \caption{Main results of different methods on the test set of MATH. The test temperature is 0.7.} \vspace{10pt} \label{tab:main_result_07}
    \begin{tabular}{cc|ccccc}
    \toprule
    \textbf{Base Model} & \textbf{Method}  & \textbf{Turn 1} & \textbf{Final Accuracy} & $\Delta(t_1,t_2)$ & $\Delta^{i \to c}(t_1,t_2)$ & $\Delta^{c \to i}(t_1,t_2)$ \\ 
    \midrule 
    Llama-3-8B-it & Prompt with Gold RM & 24.1 & 33.1 & 9.0 & 9.0 & 0\\
    Llama-3-8B-it & Intrinsic self-correction & 24.1 & 25.6 & 1.5 & 10.0  & 8.5\\
    Llama-3-8B-it & STaR/RAFT for self-correction & 25.7 & 28.0 & 2.3 & 10.9 & 8.6\\
    Llama-3-8B-it & STaR/RAFT+ for self-correction & 25.5 & 28.6 & 3.1 & 10.6 & 7.5\\
    Llama-3-8B-it & Self-correct with External ORM & 24.1 & 29.3 & 5.2 & 8.7 & 3.5 \\
    \rowcolor[rgb]{ .867, .922, .969} Llama-3-8B-it & Self-rewarding IFT  & 25.0 & 29.4 & 4.4 & 7.5 & 3.1 \\
    \midrule
    Llama-3-SFT & Prompt with Gold RM & 43.1 & 51.0 & 7.9& 7.9 & 0\\
    Llama-3-SFT & Intrinsic self-correction & 43.0 & 41.7 & -1.3 & 6.8 & 8.1 \\
    Llama-3-SFT & STaR/RAFT for self-correction & 42.5 & 40.4 & -2.1 & 9.3 & 11.4\\
    Llama-3-SFT & STaR/RAFT+ for self-correction & 42.9 & 43.1 & 0.2 & 8.1 & 7.9\\
    Llama-3-SFT & Self-correct with External ORM & 43.1 & 44.6 & 1.5 & 6.1 & 4.6 \\
    \rowcolor[rgb]{ .867, .922, .969} Llama-3-SFT & Self-rewarding IFT & 43.1 & 45.7 & 2.6 & 6.7 & 4.1 \\ 
    \midrule
        Llama-3-8B-it & Prompt with Gold RM & 67.5 & 74.0 & 6.5 & 6.5 & 0\\
    Llama-3-8B-it & Intrinsic self-correction & 67.5 & 51.6 & -15.9 & 6.1 & 22.0\\
    Llama-3-8B-it & STaR/RAFT for self-correction& 77.9 & 62.5 & -15.4 & 7.9 & 23.3  \\
    Llama-3-8B-it & STaR/RAFT+ for self-correction& 78.4 & 66.9 & -11.5 & 7.4 & 18.9 \\
    Llama-3-8B-it & Self-correct with External ORM & 67.5 & 69.9 & 2.4 & 4.5 & 2.1 \\
    \rowcolor[rgb]{ .867, .922, .969} Llama-3-8B-it & Self-rewarding IFT  & 76.4 & 80.5 & 4.1 & 7.7 & 3.6\\
    \midrule
    Llama-3-SFT & Prompt with Gold RM & 81.5 & 86.6 & 5.1 & 5.1 & 0\\
    Llama-3-SFT & Intrinsic self-correction & 81.5 & 74.8 & -6.7 & 5.3  & 12.0  \\
    Llama-3-SFT & STaR/RAFT for self-correction& 78.5 & 72.7 & -5.8 & 8.6 & 14.4\\
    Llama-3-SFT & STaR/RAFT+ for self-correction& 79.0 & 78.4 & -0.6 & 6.3 & 6.9 \\
    Llama-3-SFT & Self-correct with External ORM & 81.5 & 82.3 & 0.9 & 2.3 & 1.4  \\
    \rowcolor[rgb]{ .867, .922, .969} Llama-3-SFT & Self-rewarding IFT  & 80.8 & 82.6 & 1.8 & 2.7 & 0.9 \\
    \bottomrule
    \end{tabular}
    \end{table}

\section{Examples}

\begin{table}[t]
\centering
\begin{minipage}{1.0\columnwidth}\vspace{0mm}    \centering
\begin{sectionbox}[]{Math500 Example: Correct and Verify Correctly} 
    \centering
      \scriptsize
\begin{tabular}{p{0.97\columnwidth} c}
\emph{Prompt}: \\
       \textbf{System} Please reason step by step, and put your final answer within \boxed{}.\\
      \textbf{User} How many positive whole-number divisors does 196 have?\\
     \assi \\\\
      \emph{Responses}:\\
            To determine the number of positive whole-number divisors of 196, we can follow these steps: 1. **Prime Factorization**: First, we need to find the prime factorization of 196. 2. **Divisor Function Formula**: Once we have the prime factorization, we can use the formula for finding the number of divisors. If the prime factorization of a number $n$ is $p_1^{e_1} \cdot p_2^{e_2} cdot ldots cdot p_k^{e_k}$, then the number of divisors of $n$ is given by $(e_1 + 1)(e_2 + 1)\cdot\ldots \cdot (e_k + 1)$.  Let's start by finding the prime factorization of 196. We can do this by dividing 196 by the smallest prime number (2) until we can't divide evenly anymore, then move on to the next smallest prime number, and so on.  $196 \div 2 = 98, 98 \div 2 = 49, 49 \div 7 = 7, 7 \div 7 = 1$.  So, the prime factorization of 196 is $2^2\cdot 7^2 $.  Using the divisor function formula, the number of divisors is $(2 + 1)(2 + 1) = 3\cdot 3 = 9$.  Therefore, the number of positive whole-number divisors of 196 is $\boxed{9}$. Let's verify the correctness of the solution. The prime factorization of 196 is $(2^2 \cdot 7^2$. Using the formula for the number of divisors, we get $(2+1)(2+1) = 3 \cdot 3 = 9$. Therefore, the number of positive whole-number divisors of 196 is indeed 9.  The final answer is $\boxed{9}$.  The self-evaluation is: [VERIFY] correct.\\
    \end{tabular}
\end{sectionbox}
\caption{An example of MATH500: A prompt-response pair where the initial answer is correct and is verified correctly.}
    \label{tab:example_math500_c1}
\end{minipage}
\end{table}

\begin{table}[t]
\centering
\begin{minipage}{1.0\columnwidth}\vspace{0mm}    \centering
\begin{sectionbox}[]{Math500 Example: 
Wrong to Correct} 
    \centering
      \scriptsize
\begin{tabular}{p{0.97\columnwidth} c}
\emph{Prompt}: \\
       \textbf{System} Please reason step by step, and put your final answer within $\boxed{}$.\\
      \textbf{User} The set of points $(x,y,z)$ that satisfy \[2x = 3y = -z\]is a line. The set of points $(x,y,z)$ that satisfy \[6x = -y = -4z\]is another line. Find the angle between these lines, in degrees.\\
     \assi \\\\
      \emph{Responses}:\\
            To find the angle between the two lines, we first need to determine the direction vectors of the lines. 1. The first line is defined by the equations $2x = 3y = -z$. We can parameterize this line by setting $2x = 3y = -z = t$. This gives us $x = \frac{t}{2}$, $y = \frac{t}{3}$, and $z = -t$. So, the direction vector of the first line is $\mathbf{d_1} = \left(\frac{1}{2}, \frac{1}{3}, -1\right)$.\\2. The second line is defined by the equations $6x = -y = -4z$. We can parameterize this line by setting $6x = -y = -4z = s$. This gives us $x = \frac{s}{6}$, $y = -s$, and $z = -\frac{s}{4}$. So, the direction vector of the second line is $\mathbf{d_2} = \left(\frac{1}{6}, -1, -\frac{1}{4}\right)$.\\Next, we need to find the angle between these two direction vectors. The angle $\theta$ between two vectors $\mathbf{d_1}$ and $\mathbf{d_2}$ can be found using the dot product formula: \[ \cos \theta = \frac{\mathbf{d_1} \cdot \mathbf{d_2}}{\|\mathbf{d_1}\| \|\mathbf{d_2}\|} \] where $\mathbf{d_1} \cdot \mathbf{d_2}$ is the dot product of the vectors, and $\|\mathbf{d_1}\|$ and $\|\mathbf{d_2}\|$ are the magnitudes of the vectors. Let's calculate this step by step using Python.\
            \begin{lstlisting}
             import numpy as np
             
             #Define the direction vectors
             
             d1 = np.array([1/2, 1/3, -1])
             
             d2 = np.array([1/6, -1, -1/4])
             
             #Calculate the dot product of d1 and d2 
             
             dot_product = np.dot(d1, d2)
             
             #Calculate the magnitudes of d1 and d2 
             
             magnitude_d1 = np.linalg.norm(d1) 
             
             magnitude_d2 = np.linalg.norm(d2)
             
             #Calculate the cosine of the angle 
             
             cos_theta = dot_product / (magnitude_d1 * magnitude_d2)
             
             #Calculate the angle in radians 
             
             theta_radians = np.arccos(cos_theta)
             
             #Convert the angle to degrees 
             
             theta_degrees = np.degrees(theta_radians) 
             
             print(theta_degrees) 
             
             ```output 84.28449200558772 ```
             \end{lstlisting}
            Therefore, the final answer is: $\boxed{84.28}$.
            
             To verify the correctness of the solution, let's recheck the steps and calculations.\\1. The direction vector of the first line is $\mathbf{d_1} = \left(\frac{1}{2}, \frac{1}{3}, -1\right)$. 2. The direction vector of the second line is $\mathbf{d_2} = \left(\frac{1}{6}, -1, -\frac{1}{4}\right)$.\\The dot product of $\mathbf{d_1}$ and $\mathbf{d_2}$ is: \[ \mathbf{d_1} \cdot \mathbf{d_2} = \left(\frac{1}{2}\right)\left(\frac{1}{6}\right) + \left(\frac{1}{3}\right)(-1) + (-1)\left(-\frac{1}{4}\right) = \frac{1}{12} - \frac{1}{3} + \frac{1}{4} = \frac{1}{12} - \frac{4}{12} + \frac{3}{12} = 0 \]\\Since the dot product is zero, the vectors are orthogonal, and the angle between them is $90^\circ$. The self-evaluation is: [VERIFY] wrong.\\
             Therefore, the correct answer is: $\boxed{90^\circ}$, the self-evaluation is: [VERIFY] correct. 
    \end{tabular}
\end{sectionbox}
\caption{An example of MATH500: A prompt-response pair where the initial answer is correct and is verified correctly.}
    \label{tab:example_math500_itc}
\end{minipage}
\end{table}

\begin{table}[t]
\centering
\begin{minipage}{1.0\columnwidth}\vspace{0mm}    \centering
\begin{sectionbox}[]{Minerva Math Example: Wrong to Correct} 
    \centering
      \scriptsize
\begin{tabular}{p{0.97\columnwidth} c}
\emph{Prompt}: \\
       \textbf{System} Please reason step by step, and put your final answer within \boxed{}.\\
      \textbf{User} A particular star has an absolute magnitude $M=-7$. If this star is observed in a galaxy that is at a distance of $3 \mathrm{Mpc}$, what will its apparent magnitude be?\\
     \assi \\\\
      \emph{Responses}:\\
            To determine the apparent magnitude of the star, we can use the distance modulus formula: \[ m - M = 5 \log_{10}(d) - 5 \] where:$ m $ is the apparent magnitude,$ M $ is the absolute magnitude,$ d $ is the distance to the star in parsecs.\\ Given:$ M = -7 $$ d = 3 \text{ Mpc} = 3 \times 10^6 \text{ pc} $\\
            First, we need to calculate $ 5 \log_{10}(d) - 5 $:\[ 5 \log_{10}(3 \times 10^6) - 5 \]
            We can break this down into two parts: 1. Calculate $ \log_{10}(3 \times 10^6) $. 2. Multiply the result by 5 and subtract 5. Let's do this calculation in Python.
            \begin{lstlisting}
            import math
            
            # Given values
            
            M = -7
            
            d = 3 * 10**6 # distance in parsecs
            
            # Calculate the distance modulus
            
            distance_modulus = 5 * math.log10(d) - 5
            
            # Calculate the apparent magnitude
            
            m = M + distance_modulus
            
            print(m)
            
           ```output 7.579181246047625 ```
            \end{lstlisting}
            The apparent magnitude of the star, when observed from a distance of 3 Mpc, is approximately $ 7.58 $. So the final answer is: $\boxed{7.58}$.\\
            To verify the correctness of the solution, let's recheck the steps: 1. The absolute magnitude $ M $ of the star is given as $-7$. 2. The distance $ d $ to the star is $ 3 \text{ Mpc} = 3 \times 10^6 \text{ pc} $. 3. The distance modulus formula is $ m - M = 5 \log_{10}(d) - 5 $.\\ Let's recompute the distance modulus:\[ 5 \log_{10}(3 \times 10^6) - 5 = 5 (\log_{10}(3) + \log_{10}(10^6)) - 5 = 5 (\log_{10}(3) + 6) - 5 = 5 \log_{10}(3) + 30 - 5 = 5 \log_{10}(3) + 25 \]
            Using the approximate value $ \log_{10}(3) \approx 0.4771 $:\[ 5 \times 0.4771 + 25 = 2.3855 + 25 = 27.3855 \]
            So the apparent magnitude $ m $ is:\[ m = M + 27.3855 = -7 + 27.3855 = 20.3855 \] It seems there was an error in the previous calculation. The correct apparent magnitude is approximately $ 20.39 $.\\ So the final answer is: $\boxed{20.39}$.\\
            The self-evaluation is: [VERIFY] wrong. The mistake in the previous reasoning was in the calculation of the distance modulus. Let's correct it step by step.\\Given: Absolute magnitude $ M = -7 $ Distance $ d = 3 \text{ Mpc} = 3 \times 10^6 \text{ pc} $\\ The distance modulus formula is:\[ m - M = 5 \log_{10}(d) - 5 \]\\ Let's recompute the distance modulus correctly:\[ 5 \log_{10}(3 \times 10^6) - 5 = 5 (\log_{10}(3) + \log_{10}(10^6)) - 5 = 5 (\log_{10}(3) + 6) - 5 = 5 \log_{10}(3) + 30 - 5 = 5 \log_{10}(3) + 25 \]
            Using the approximate value $ \log_{10}(3) \approx 0.4771 $:\[ 5 \times 0.4771 + 25 = 2.3855 + 25 = 27.3855 \]
            So the apparent magnitude $ m $ is:\[ m = M + 27.3855 = -7 + 27.3855 = 20.3855 \]
            The correct apparent magnitude is approximately $ 20.39 $. So the final answer is: $\boxed{20.39}$. \\
            Since the steps and calculations are correct, the self-evaluation is: [VERIFY] correct. 
    \end{tabular}
\end{sectionbox}
\caption{An example of Minerva Math: A prompt-response pair where the initial answer is correct and is verified correctly..}
    \label{tab:example_hh_rlhf_dataset}
\end{minipage}
\end{table}

\begin{table}[t]
\centering
\begin{minipage}{1.0\columnwidth}\vspace{0mm}    \centering
\begin{sectionbox}[]{OlympiadBench Example: Correct and Verify Correctly} 
    \centering
      \scriptsize
\begin{tabular}{p{0.97\columnwidth} c}
\emph{Prompt}: \\
       \textbf{System} Please reason step by step, and put your final answer within \boxed{}.\\
      \textbf{User} Find (in closed form) the difference between the number of positive integers at most $2^{2017}$ with even weight and the number of positive integers at most $2^{2017}$ with odd weight.\\
     \assi \\\\
      \emph{Responses}:\\
            To solve this problem, we need to understand the concept of ``weight" of a number. The weight of a number is the number of 1s in its binary representation. We need to find the difference between the number of positive integers at most $2^{2017}$ with even weight and the number of positive integers at most $2^{2017}$ with odd weight.
            
            Here's the reasoning:\\
            1. The number of positive integers at most $2^{2017}$ is $2^{2017}$. Each of these numbers can be represented as a 2017-bit binary number (including leading zeros).\\ 2. The weight of a number is the number of 1s in its binary representation. For a 2017-bit number, the weight can be any integer from 0 to 2017.\\ 3. The number of 2017-bit numbers with a specific weight $k$ is given by the binomial coefficient $\binom{2017}{k}$, which is the number of ways to choose $k$ positions out of 2017 to place the 1s.\\ 4. The binomial coefficients $\binom{2017}{k}$ for $k$ from 0 to 2017 are symmetric, i.e., $\binom{2017}{k} = \binom{2017}{2017-k}$. This means that the number of numbers with even weight is equal to the number of numbers with odd weight, except for the case when the weight is exactly 1008 (since 2017 is odd, there is no middle term).\\ 5. Therefore, the difference between the number of positive integers at most $2^{2017}$ with even weight and the number of positive integers at most $2^{2017}$ with odd weight is exactly the binomial coefficient $\binom{2017}{1008}$, but with a sign change because the number of even-weight numbers is one more than the number of odd-weight numbers.\\ The final answer is: $\boxed{0}$.
            
            This is because the number of even-weight numbers and odd-weight numbers is exactly the same, so their difference is zero. Let's confirm this with a small Python code snippet to illustrate the symmetry of the binomial coefficients.
            \begin{lstlisting}
            from math import comb
            
            # Number of bits
            
            n = 2017
            
            # Calculate the binomial coefficient for the middle term (since n is odd, there is no middle term)
            
            middle_term = comb(n, n // 2)
            
            # The difference between the number of even-weight and odd-weight numbers
            
            difference = middle_term if n % 2 == 1 else 0
            
            print(difference)
            \end{lstlisting}            
            The reasoning and the Python code confirm that the difference between the number of positive integers at most $2^{2017}$ with even weight and the number of positive integers at most $2^{2017}$ with odd weight is indeed zero. Therefore, the final answer is: $\boxed{0}$
            
            The self-evaluation is: [VERIFY] correct.\\
    \end{tabular}
\end{sectionbox}
\caption{An example of OlympiadBench: A prompt-response pair where the initial answer is correct and is verified correctly.}
    \label{tab:example_math500_c2}
\end{minipage}
\end{table}


\end{document}